%% file: arxiv_version.tex
\documentclass{article}

\usepackage[preprint,nonatbib]{neurips_2025}

\usepackage{cite}
\usepackage[table]{xcolor}

\usepackage{enumitem}
\usepackage{tabularx}       
\usepackage{array}          
\newcolumntype{Y}{>{\raggedright\arraybackslash}X}

\usepackage{makecell}
\usepackage{multirow} 
\usepackage{multicol}
\usepackage{amsmath}

\usepackage{booktabs}

\usepackage{ragged2e} 
\usepackage{geometry}
\usepackage{tcolorbox}
\tcbuselibrary{skins,breakable}
\usepackage[table]{xcolor}
\definecolor{lightgray}{gray}{0.9}

\usepackage[utf8]{inputenc} 
\usepackage[T1]{fontenc}    
\usepackage{hyperref}       
\usepackage{url}            
\usepackage{booktabs}       
\usepackage{amsfonts}       
\usepackage{nicefrac}       
\usepackage{microtype}      
\usepackage{xcolor}         
\usepackage{graphicx} 
\usepackage{hhline}

\usepackage{graphicx}   
\usepackage{adjustbox}
\usepackage{subcaption} 
\usepackage{etoc}    
\usepackage{hyperref}     

\usepackage{graphicx}    
\usepackage[abs]{overpic} 
\usepackage{rotating}  
\usepackage{tcolorbox}                  
\usepackage{enumitem}                   

\title{Interpreting Chest X-rays Like a Radiologist: A Benchmark with Clinical Reasoning}

\setlist[description]{
  style=nextline,       
  labelwidth=3.5em,     
  labelsep=0.5em,       
  leftmargin=!,         
  itemsep=0.5ex,        
  font=\normalfont\bfseries  
}

\setlist[itemize]{
  leftmargin=*,         
  itemsep=0.3ex         
}

\begin{document}

\author{
  \textbf{Jinquan Guan\textsuperscript{1}}, %
 \textbf{Qi Chen\textsuperscript{3}}, %
 \textbf{Lizhou Liang\textsuperscript{1}}, %
  \textbf{Yuhang Liu\textsuperscript{1}},  
 \textbf{Vu Minh Hieu Phan\textsuperscript{3}}, \and
 \textbf{Minh-Son To\textsuperscript{4}}, %
 \textbf{Jian Chen\textsuperscript{1}}, %
 \textbf{Yutong Xie\textsuperscript{2}} \\ 
 \\
 \small
  \textsuperscript{1} Software Engneering, South China University of Technology \\
 \small
  \textsuperscript{2} Department of Computer Vision, MBZUAI \\
 \small
  \textsuperscript{3} Australia Institute of Machine Learning, University of Adelaide \\
 \small
  \textsuperscript{4} Flinders Health and Medical Research Institute, Flinders University \\
}

\etocdepthtag.toc{mtchapter}

\maketitle

\begin{abstract}

%

Artificial intelligence (AI)-based chest X-ray (CXR) interpretation assistants have demonstrated significant progress and are increasingly being applied in clinical settings. However, contemporary medical AI models often adhere to a simplistic input-to-output paradigm, directly processing an image and an instruction to generate a result, where the instructions may be integral to the model's architecture. This approach overlooks the modeling of the inherent diagnostic reasoning in chest X-ray interpretation. Such reasoning is typically sequential, where each interpretive stage considers the images, the current task, and the contextual information from previous stages. This oversight leads to several shortcomings, including misalignment with clinical scenarios, contextless reasoning, and untraceable errors.
To fill this gap, we construct \textbf{CXRTrek}, a new multi-stage visual question answering (VQA) dataset for CXR interpretation. The dataset is designed to explicitly simulate the diagnostic reasoning process employed by radiologists in real-world clinical settings for the first time. CXRTrek covers 8 sequential diagnostic stages, comprising 428,966 samples and over 11 million question–answer (Q\&A) pairs, with an average of 26.29 Q\&A pairs per sample.
Building on the CXRTrek dataset, we propose a new vision-language large model (VLLM), \textbf{CXRTrekNet}, specifically designed to incorporate the clinical reasoning flow into the VLLM framework. CXRTrekNet effectively models the dependencies between diagnostic stages and captures reasoning patterns within the radiological context. Trained on our dataset, the model consistently outperforms existing medical VLLMs on the CXRTrek benchmarks and demonstrates superior generalization across multiple tasks on five diverse external datasets.
The dataset and model can be found in our repository\footnote{https://github.com/guanjinquan/CXRTrek}.

\end{abstract}

\section{Introduction}
\label{sec:introduction}

Chest X-ray (CXR) is one of the most commonly used tools for body screening, playing a key role in the screening, diagnosis, and treatment monitoring of pulmonary diseases \cite{cid2024development, hirsch2001early}. 
With the rapid advancement of deep learning techniques, numerous CXR-based interpretation methods have emerged, significantly furthering research in tasks such as abnormality detection, disease identification, and report generation~\cite{ccalli2021deep,ait2023review,qin2019using}. 
These systems have shown the potential to support radiologists by improving diagnostic efficiency.


Early methods typically employed one-step, task-independent paradigms, directly mapping medical images to diagnostic labels or textual reports~\cite{rajpurkar2017chexnet,smit2020chexbert,tiu2022expert,wang2022medclip,chen2020generating,chen2024act,wang2023r2gengpt}. Although effective for narrowly defined tasks, such approaches largely overlooked the intermediate steps of clinical reasoning and the relationships among different task, thereby limiting their clinical reasoning and generalization capabilities.
Recent advances in large language models (LLMs) have motivated their integration into medical image analysis to improve clinical reasoning and generalization, such as XrayGPT~\cite{thawkar2023xraygpt}, LLaVA-Med~\cite{li2023llava}, CheXagent~\cite{smit2020chexbert}, and RadVLM~\cite{deperrois2025radvlm}. However, these models often remain restricted to simplistic task formulations, shallow reasoning chains.
Consequently, they fail to explicitly model the complex reasoning process that underlies real-world clinical diagnosis, thereby facing critical limitations in practical deployment.

Clinically, radiologists typically interpret CXR images using a logical reasoning process~\cite{klein2019systematic,gelaw2015screening,martensen2013radiographic
}. It begins with assessing image quality, followed by identifying any anomalies. If present, these anomalies are analyzed for morphological characteristics. Subsequent stages include comparing with prior images, predicting disease progression, and finally providing diagnostic conclusions and generating a report. 
The determinations at each stage depend not only on CXR images, but also on the diagnostic context and conclusions from antecedent stages. This traceable reasoning is essential for effective clinical decision-making, but current models are struggling to replicate it.


\begin{figure*}[t!]
  \centering
   \includegraphics[width=\textwidth]{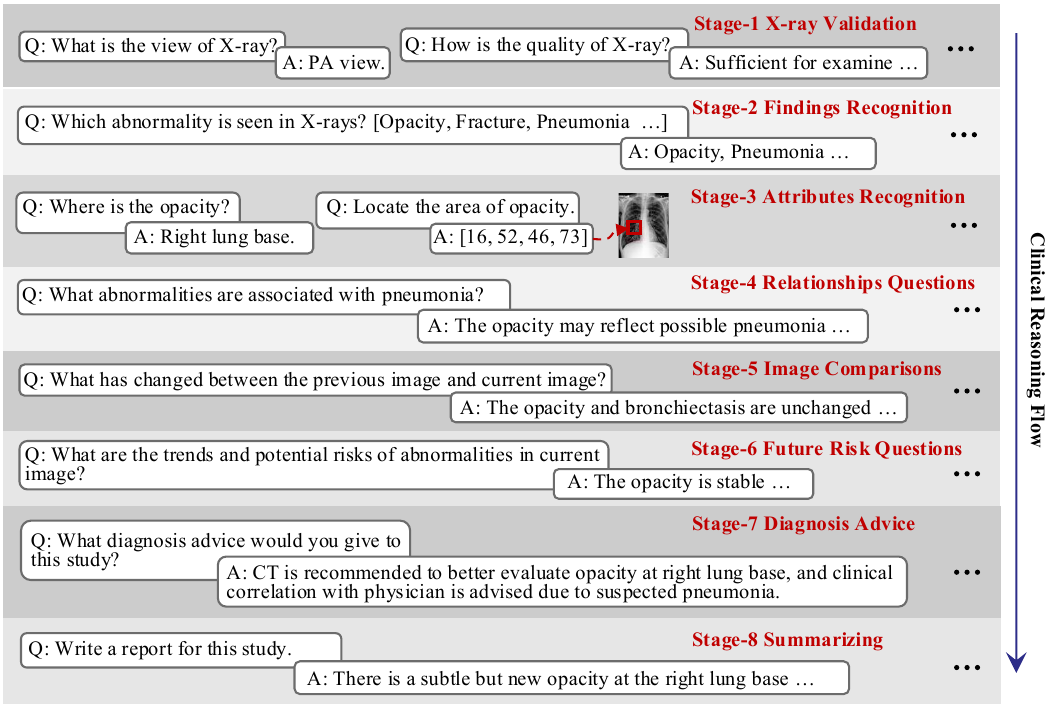}


\caption{The clinical reasoning flow in CXRTrek through a multi-stage Q\&A flow. Each Q\&A pair represents a task for a specific stage of clinical CXRs interpretation, where each stage may involve multiple Q\&A pairs for a thorough analysis of its designated responsibilities. The stages progress from image validation to report summarization, simulating the sequential reasoning of radiologists.}
  \label{fig:dataset_case}
\end{figure*}

To bridge this gap, we construct CXRTrek (see Figure~\ref{fig:dataset_case}), a new benchmark that guides models through an 8-stage clinical reasoning process for chest X-ray interpretation, simulating how radiologists progressively analyze and synthesize diagnostic information, including: (1) \textbf{Stage-1 X‑ray Validation}, which assesses the view orientation and overall image quality to ensure the validity of all subsequent analyses; (2) \textbf{Stage-2 Findings Recognition}, which identifies any abnormalities or foreign bodies in the image; (3) \textbf{Stage-3 Findings’ Attributes Recognition}, which characterizes each detected finding in detail—its severity, exact location, morphological type, and other key descriptive features; (4) \textbf{Stage-4 Findings’ Relationships Questions}, which models the spatial, causal, or semantic relationships among multiple findings within the image; (5) \textbf{Stage-5 Image Comparisons}, which compares the current image with previous studies to highlight interval changes and assess lesion evolution; (6) \textbf{Stage-6 Future Risk Questions}, which predicts the future trajectory of identified abnormalities to estimate any associated health risks; (7) \textbf{Stage-7 Diagnosis Advice}, which offers preliminary clinical recommendations to the referring physician based on the integrated analysis; and (8) \textbf{Stage-8 Summarizing}, which writes a complete diagnostic report for summarization.

Based on this dataset, we propose CXRTrekNet, a clinical reasoning-aware model tailored to the structure of CXRTrek dataset. CXRTrekNet is built on a vision large language model (VLLM) backbone and incorporates stage-specific instructions to guide the model through each phase of the diagnostic workflow. It enables information flow across stages, allowing the model to accumulate contextual knowledge and perform step-by-step reasoning aligned with clinical logic. By explicitly modeling interdependent reasoning steps, CXRTrekNet improves understanding of complex clinical semantics and supports traceable decision-making across diverse tasks.

The primary contributions of this work are threefold:
\begin{itemize}
\item We construct CXRTrek, a new benchmark that systematically decomposes CXR interpretation into 8 clinically guided stages, enabling fine-grained supervision, progressive reasoning, and multi-stage evaluation aligned with real-world radiological workflows.
\item We propose CXRTrekNet, a clinical reasoning-aware VLLM model, designed to align with real-world clinical scenarios for greater interpretability, and to enhance its in-context reasoning and semantic understanding via knowledge injection from the CXRTrek dataset.
\item We conduct comprehensive experiments across our benchmark and five clinical datasets, demonstrating the effectiveness and generalizability of CXRTrekNet across classification, detection, visual question answering, and report generation.
\end{itemize}

\section{Related work}
\label{sec:related_work}




\paragraph{Chest X-ray interpretation methods.}
CXR interpretation, a key medical AI research area, has seen significant progress in image classification, report generation, and visual-language modeling (VLM). Early methods, including CheXNet~\cite{rajpurkar2017chexnet} and CheXbert~\cite{smit2020chexbert}, utilized deep learning for multi-label classification to identify diseases or abnormalities in CXRs. Subsequent research expanded to automated report generation, with models such as R2GenGPT~\cite{wang2023r2gengpt}, TieNet~\cite{wang2018tienet}, and CVT2DistilGPT2~\cite{nicolson_improving_2023} designed to produce textual diagnostic reports from images. Furthermore, advanced visual-language models like GLoRIA~\cite{huang2021gloria}, MedCLIP~\cite{wang2022medclip}, CheXzero~\cite{tiu2022expert}, ConVIRT~\cite{zhang2022contrastive}, and PubMedCLIP~\cite{eslami2023pubmedclip} enhance semantic consistency and cross-modal learning via textual supervision and multimodal alignment.
However, these models often use a single-step mapping strategy, overlooking the multi-step reasoning crucial in clinical practice. Although models like Rad-Restruct's~\cite{pellegrini2023rad} hierarchical Q\&A model aim for step-by-step analysis, they are often limited by simplistic questions and poor clinical alignment. This limitation leads to reduced trustworthiness and limited interpretability, and hinders their capacity to handle complex diagnostic tasks that require step-by-step reasoning.

\paragraph{Medical visual large language models (VLLMs).}  
VLLMs have demonstrated strong performance on a variety of vision–language tasks. However, general-purpose reasoning models such as GPT-4~\cite{achiam2023gpt} focus primarily on mathematical problem solving, code generation, and general conversational fluency, which limits their effectiveness for specialized medical inference. Medical VLLMs designed specifically for medical use—such as XrayGPT\cite{thawkar2023xraygpt}, LLaVA-Med\cite{li2023llava}, MiniGPT-Med~\cite{alkhaldi2024minigpt}, CheXagent~\cite{chen2024chexagent}, BioMedGPT\cite{luo2023biomedgpt}, RadFM~\cite{wu2023towards}, MedVINT~\cite{zhang2023pmc}, HuaTuoGPT-Vision~\cite{chen2024huatuogpt} and RadVLM~\cite{deperrois2025radvlm}, excel at handling medical terminology and aligning visual and textual information. Nevertheless, these models lack access to explicit multi-step reasoning examples and are constrained by short context windows, which restricts their ability to perform deep, context-aware clinical reasoning.  
\emph{To bridge these gaps, we generate multi-stage diagnostic Q\&A and leverage them to train a novel medical VLLM, equipping it with context-aware reasoning capabilities for clinical scenarios.}

\section{CXRTrek dataset construction}
\label{sec:dataset_construction}




\subsection{Dataset design principles}

CXRTrek is constructed from two large-scale datasets, MIMIC-CXR~\cite{johnson2019mimic} and CheXpert-plus~\cite{chambon2024chexpert}, which paired CXRs with radiology reports. To design a dataset that facilitates clinical reasoning and reflects real-world scenarios, we investigated existing knowledge and standards in human CXR interpretation. Based on this comprehensive investigation, we identified 8 core diagnostic stages for CXR interpretation and structured their sequence to reflect the clinical reasoning flow.

\paragraph{Rationale for 8 stages and their clinical reasoning flow.}
To emulate the clinical reasoning process of radiologists, we reviewed established radiology guidelines~\cite{klein2019systematic,gelaw2015screening,martensen2013radiographic} and consulted with domain experts. Building on this foundation, we define 8 progressive interpretation stages that reflect the core diagnostic workflow, as illustrated in Figure~\ref{fig:dataset_case}. We found supporting evidence in established guidelines affirming that these 8 stages are both core and necessary. For instance, \textbf{Stage-1 (X-ray Validation)} is directly informed by Klein \textit{et al.}~\cite{klein2019systematic}, who emphasize that ensuring the technical adequacy of chest radiographs is critical to prevent diagnostic errors and ensure reliable subsequent analyses. Furthermore, \textbf{Stage-3 (Attributes Recognition)}, following \textbf{Stage-2 (Findings Recognition)}, implements Martensen \textit{et al.}~\cite{martensen2013radiographic}'s recommendation to assess key abnormality features ($e.g.$, morphology, density) in detail, which is essential for accurate diagnostic reasoning.

Intuitively, the formation of a clinical reasoning flow is logical. For example, \textbf{Stage-8 (Summarizing)}, which involves generating a comprehensive report, naturally relies on the analyses performed in the preceding stages. Similarly, \textbf{Stage-4 (Relationships Questions)}, focusing on the interplay between different findings, must be based on the findings identified in earlier stages.

Further details on the guideline-based evidence supporting the design of this 8-stage flow, and how each stage aligns with clinical scenarios, are provided in the Appendix Section B.


\subsection{Question and answer generation}

\paragraph{Question generation.}

We surveyed 16 public datasets~\cite{irvin2019chexpert,nguyen2022vindr,pellegrini2023rad,ben2019vqa,lau2018dataset,bae2024mimic,liu2021slake,hu2023expert,wu2021chest,bustos2020padchest,national2011national,bannur2023ms,boecking2022ms,humedical,demner2016preparing,rsna-pneumonia-detection-challenge}, cataloging their questions and clinical objectives. The identified questions cover a wide range of tasks, including abnormality detection, view recognition, disease progression analysis, report generation, phrase grounding, and more. These questions are aligned with our 8-stage interpretation flow. We supplemented this by using GPT-4~\cite{achiam2023gpt} to generate additional question templates, then manually refining them into the final question set. More details on these questions are provided in Appendix Section B.

\paragraph{Answer generation.}

To construct high-quality supervision for all 8 stages of CXRTrek, we design a hybrid annotation approach that generates answer for each question using a combination of rule-based approaches, LLM outputs, and annotations from other datasets~\cite{wu2021chest,johnson2019mimic,chambon2024chexpert}.

We begin by identifying clinically significant entities—specifically abnormalities and foreign bodies—along with their associated attributes. We use the LLaMA 3.2 8B~\cite{grattafiori2024llama} as the tool to extract this information. For abnormalities, we extract four attributes: \textit{location}, \textit{trait} ($e.g.,$ scattered, chronic), \textit{severity} ($e.g.,$ mild, severe), and \textit{trend} ($e.g.,$ improving). For foreign bodies, we extract \textit{location} and \textit{trait}. To reduce hallucinations, we construct a domain-specific medical vocabulary by aggregating terms from UMLS~\cite{bodenreider2004unified} and nine public datasets~\cite{liu2021slake,ben2019vqa,lau2018dataset,pellegrini2023rad,liu2024medcot,hu2023expert,bustos2020padchest,wu2021chest}, followed by expert refinement to produce a final list of over 5,000 clinically relevant terms. An extracted entity is retained only if at least 50\% of its tokens match this vocabulary.

In addition to entity-level information, we extract 3 types of diagnostic context embedded in narrative form. Specifically, we identify: (1) \textit{X-ray quality assessment}, which evaluates the quality of the X-ray; (2) \textit{findings' relationships}, which captures causal or spatial dependencies among findings; and (3) \textit{diagnosis advice}, such as follow-up or treatment advice. These cues are not explicitly labeled in previous datasets, so we employ LLM-based extraction with customized prompts to identify them.

Based on the extracted information, answers for questions in each stage are generated using a hybrid approach that combines LLM outputs, predefined rules, and annotations from external datasets~\cite{wu2021chest,johnson2019mimic,chambon2024chexpert}:
\textbf{Stage-1 (X-ray Validation)} utilizes X-ray view information from orignal dataset and LLM-extracted quality assessments;
\textbf{Stage-2 (Findings Recognition)} derives answers from LLM-extracted entities;
\textbf{Stage-3 (Attributes Recognition)} leverages LLM-extracted attributes, supplemented by Chest ImaGenome~\cite{wu2021chest} annotations for detecting findings or anatomical structures;
\textbf{Stage-4 (Relationship Questions)} relies on LLM-extracted findings' relationships;
\textbf{Stage-5 (Image Comparisons)} and \textbf{Stage-6 (Future Risk Questions)} apply rule-based methods from MIMIC-Diff-VQA~\cite{hu2023expert} to compare current/previous entities and current/subsequent studies, respectively to generate answer;
\textbf{Stage-7 (Diagnosis Advice)} generates answers from LLM-extracted diagnostic advice; and
finally, \textbf{Stage-8 (Summarizing)} employs the original X-ray(s) reports as answer.

Finally, our dataset utilizes 4 input-output formats for question answering: (1) \textit{Open-ended}, requiring free-form text; (2) \textit{Closed-ended}, limited to Yes or No; (3) \textit{Choice}, including single- and multi-choice options; and (4) \textit{Detection}, localizing with bounding boxes.
Inter-task dependencies are encoded as a directed acyclic graph, enforcing logical consistency and reflecting the stepwise nature of clinical workflows.
More details are provided in the Appendix Section B.


\subsection{Dataset statistics and annotation quality analysis}

\begin{figure*}[t!]
  \centering

  \begin{minipage}{\textwidth}
    \centering
    \begin{subfigure}[t]{0.323\textwidth}
      \centering
      \includegraphics[width=\linewidth]{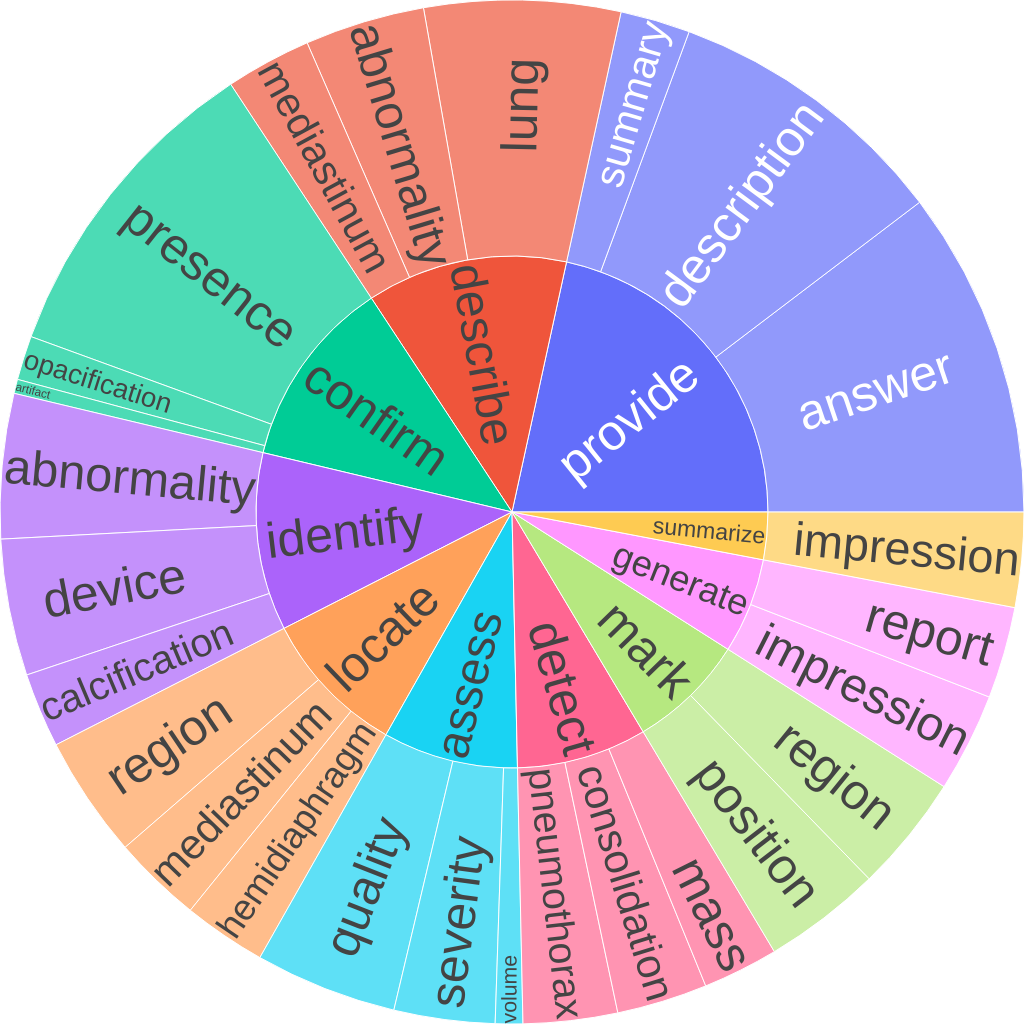}
      \caption{Instruction Word‑Pairs}
      \label{fig:subfig1}
    \end{subfigure}\hfill%
    \begin{subfigure}[t]{0.323\textwidth}
      \centering
      \includegraphics[width=\linewidth]{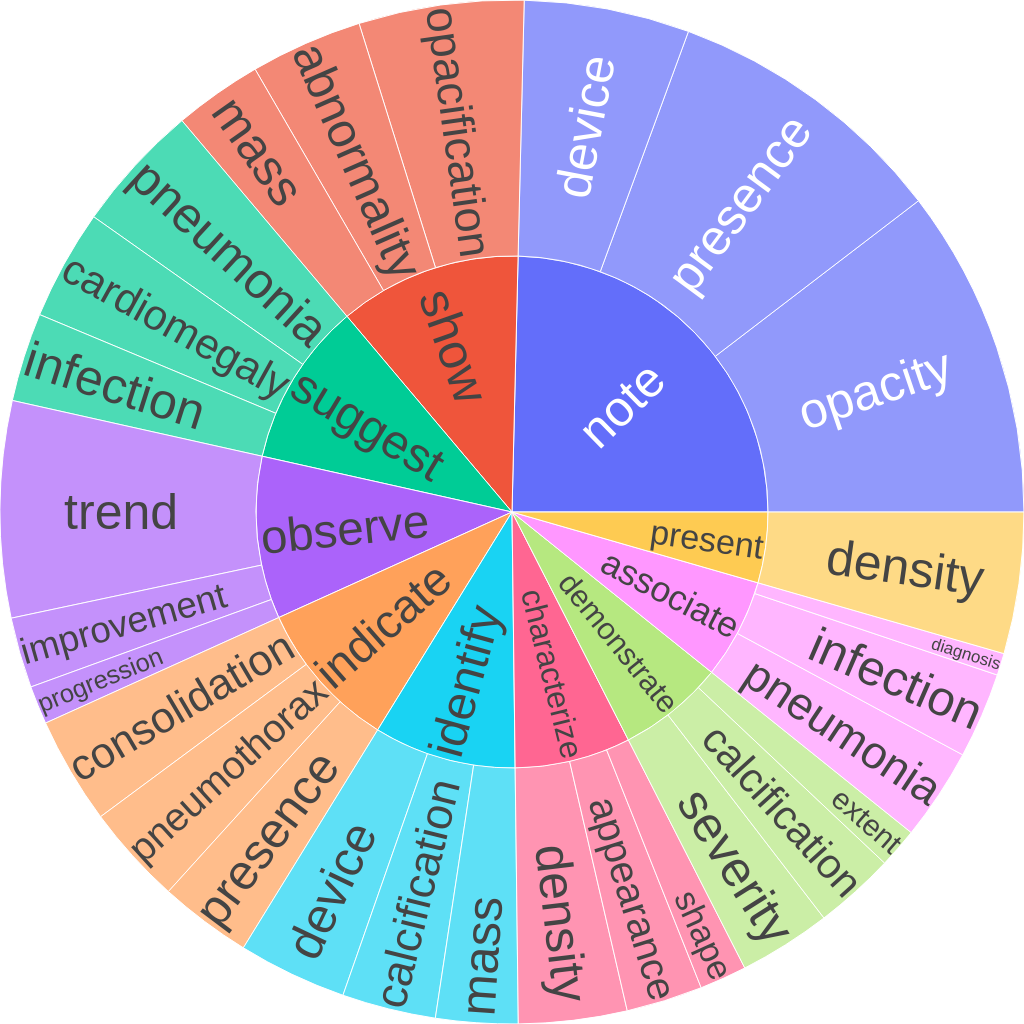}
      \caption{Response Word‑Pairs}
      \label{fig:subfig2}
    \end{subfigure}\hfill%
    \begin{subfigure}[t]{0.323\textwidth}
      \centering
      \includegraphics[width=\linewidth]{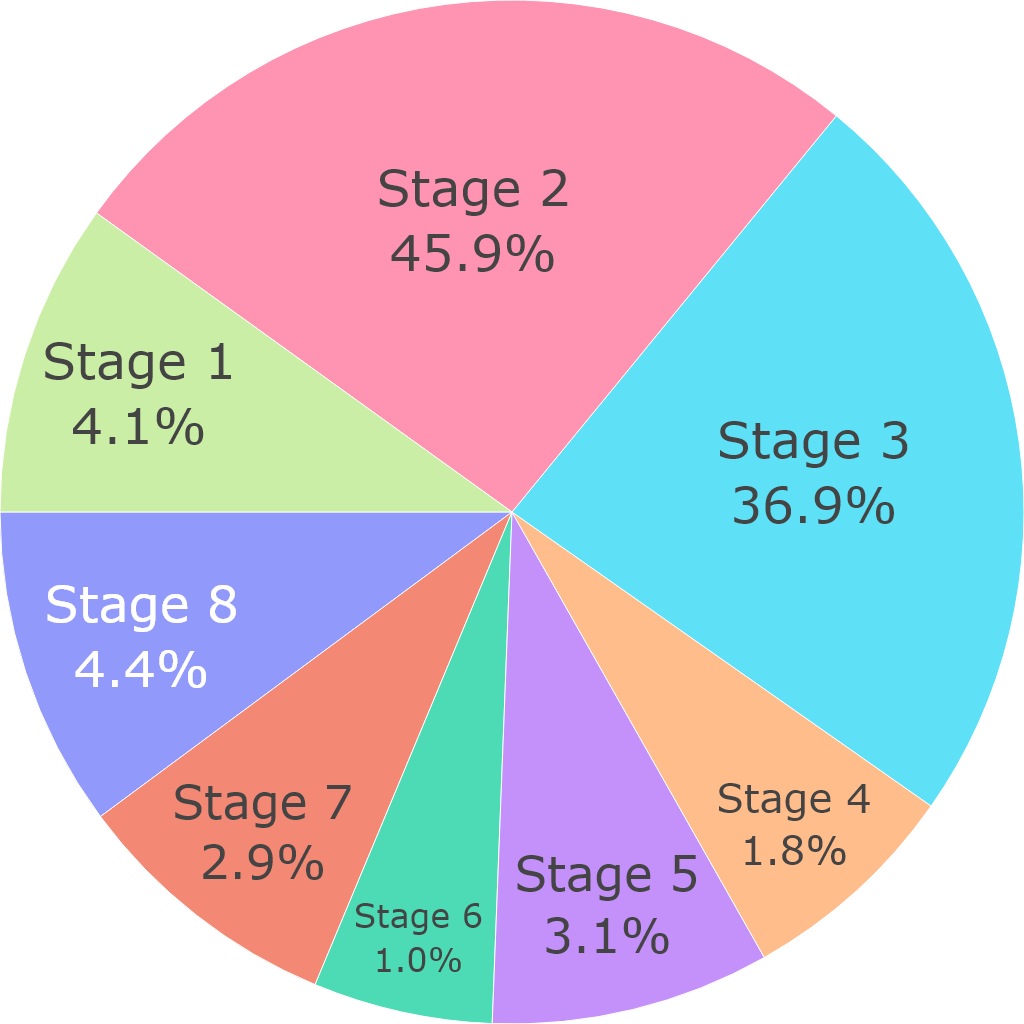}
      \caption{Number of Q\&A pairs.}
      \label{fig:subfig3}
    \end{subfigure}
  \end{minipage}

  \vspace{1ex}

  \begin{minipage}{\textwidth}
  \centering
  \makebox[0pt][r]{%
  \raisebox{+0.35cm}{
    \rotatebox[origin=c]{90}{\scriptsize Normalized Frequency}%
  }%
}%
\hspace{0.1pt}
  \begin{subfigure}[c]{0.322\textwidth}
    \adjustbox{max height=3cm,valign=c}{%
      \includegraphics[width=\linewidth]{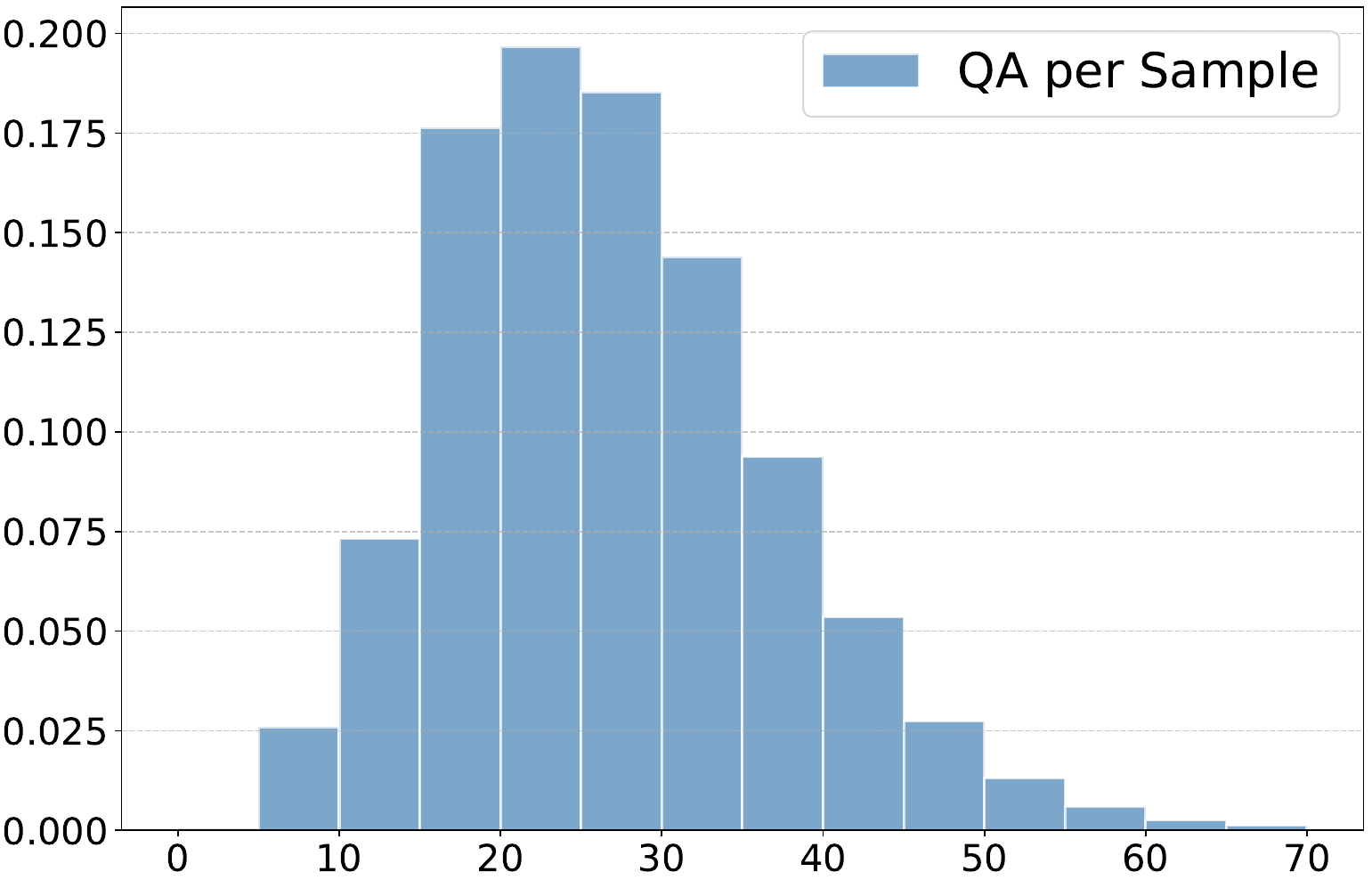}
    }
    \caption{Q\&A per Sample Distribution}
  \end{subfigure}\hfill%
  \begin{subfigure}[c]{0.326\textwidth}
    \adjustbox{max height=3cm,valign=c}{%
      \includegraphics[width=\linewidth]{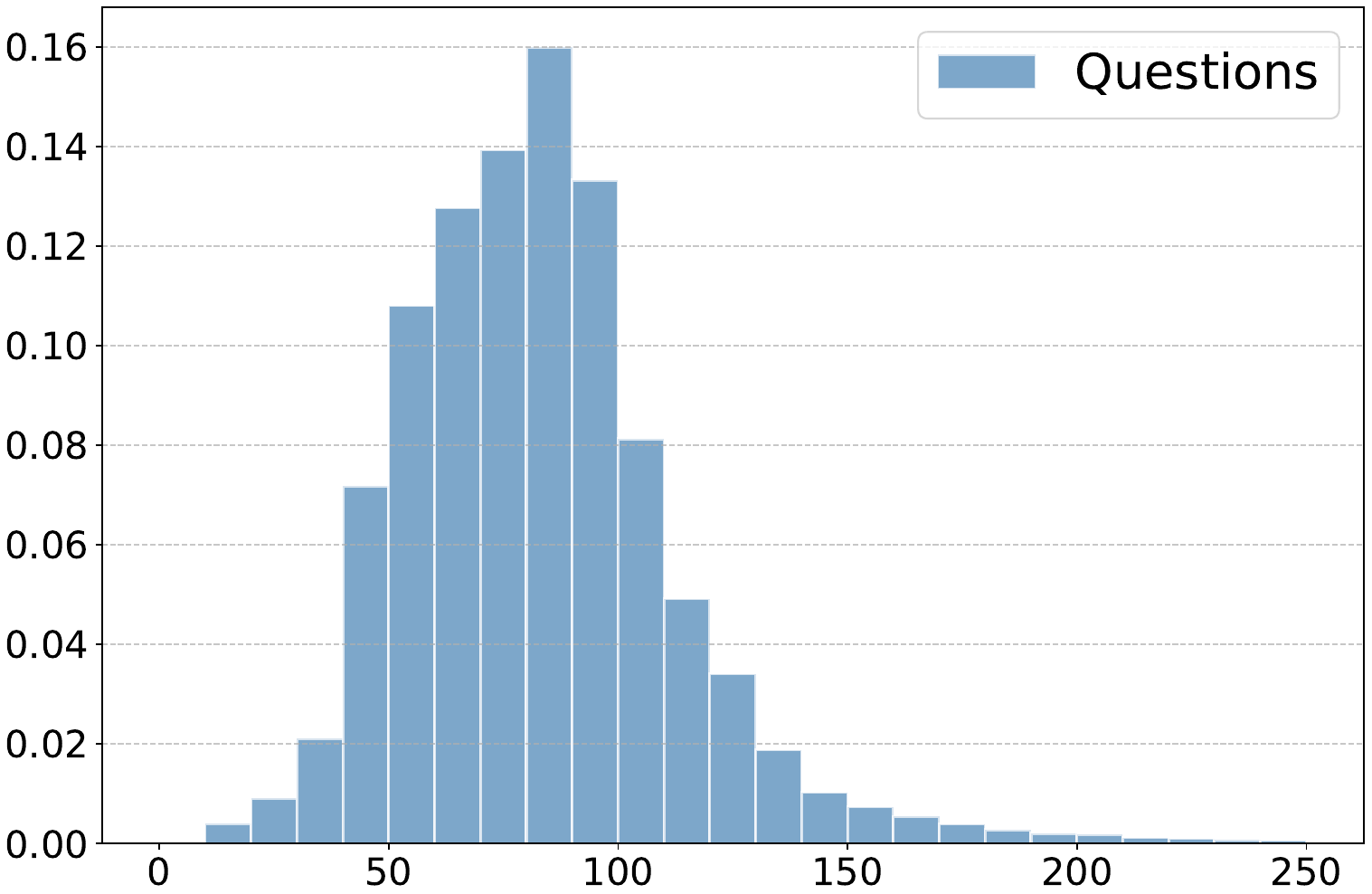}
    }
    \caption{Question Length Distribution}
  \end{subfigure}\hfill%
  \begin{subfigure}[c]{0.326\textwidth}
    \adjustbox{max height=3cm,valign=c}{%
      \includegraphics[width=\linewidth]{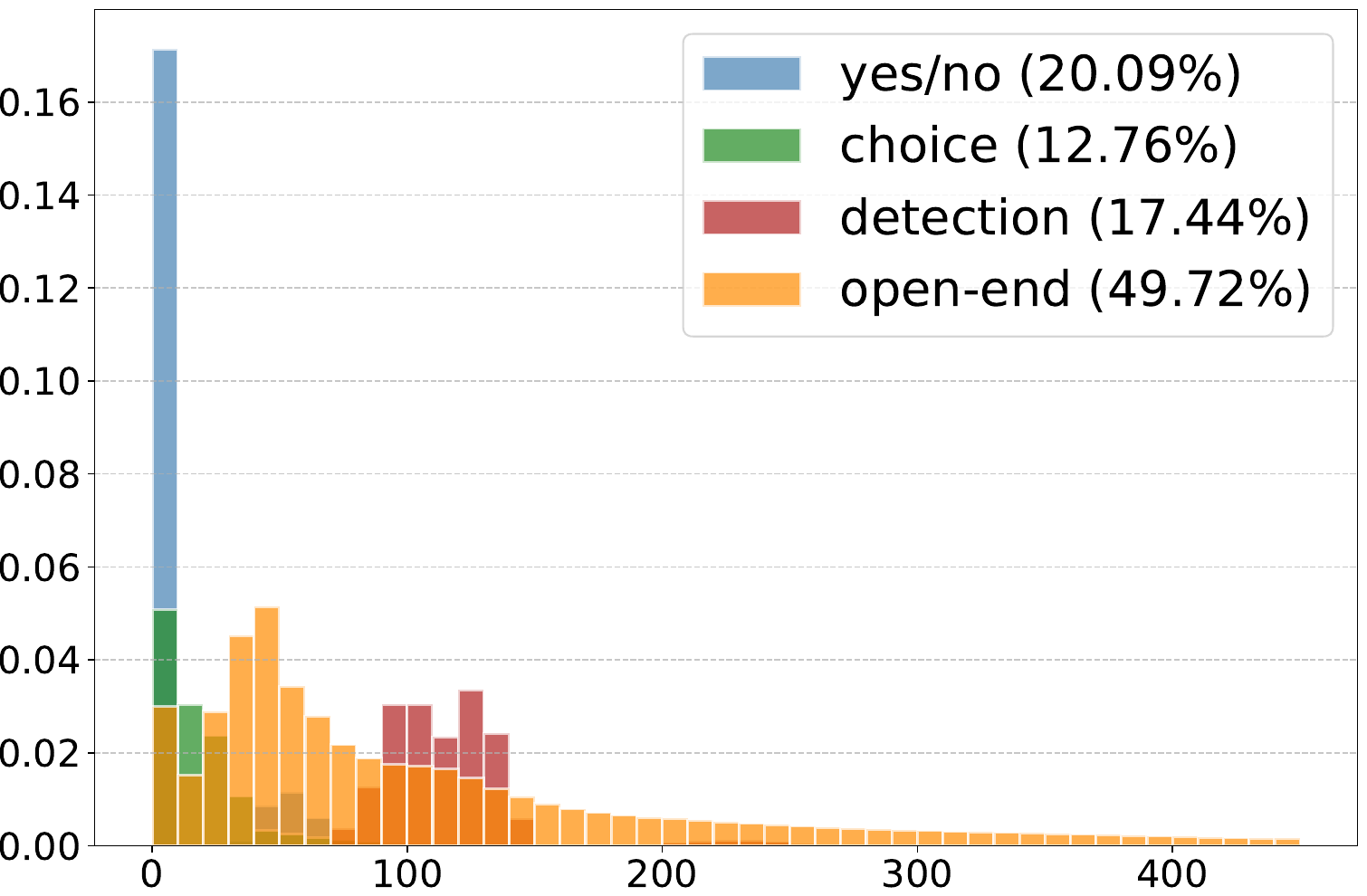}
    }
    \caption{Answer Length Distribution}
  \end{subfigure}
\end{minipage}

  \caption{Frequency analysis of word-pairs and task distributions in the training set of CXRTrek. 
  (a) Instruction word-pairs generated using GPT-4\cite{achiam2023gpt} highlight dominant terms in questions  ($e.g.$, "identify") . 
  (b) Response word-pairs show common diagnostic observations ($e.g.$, "present", "opacity"). 
  (c) Distribution of the number of Q\&A pairs across 8 stages.
  (d)-(f) Distribution of Q\&A pairs per sample, question/answer length statistics, and proportions of the four response formats in CXRTrek. Normalized frequency is the item count divided by the total count across all items.}
  \label{fig:training_set_statistic}
\end{figure*}

The CXRTrek training set is constructed from the MIMIC-CXR~\cite{johnson2019mimic} training split and the CheXpert-plus~\cite{chambon2024chexpert} public dataset. It covers 8 diagnostic stages, comprising 428,966 CXRs and over 11 million VQA pairs, with an average of 26.29 Q\&A pairs per sample. The statistics of CXRTrek training set are visualized in Figure~\ref{fig:training_set_statistic}, demonstrating the distribution characteristics across different stages and Q\&As.
To validate the quality of the automatically extracted labels, we randomly select 100 samples from the training set. For each sample, human annotators manually extract all labels targeted by the model. We then compare the model-extracted outputs with human references and achieve a high BERTScore~\cite{zhang2019bertscore} of 92.53, validating the reliability of our annotation extraction framework.

In addition, we select 200 samples from the MIMIC-CXR~\cite{johnson2019mimic} test set to construct the CXRTrek test set and perform human verification to ensure annotation quality for model evaluation.

%



\section{CXRTrekNet: a clinical reasoning-aware VLLM}
\label{sec:method}

CXRTrekNet is a VLLM tailored to simulate the structured diagnostic reasoning process commonly adopted by radiologists.
Specifically, given one or more chest X-ray images and a sequential set of clinically guided questions, the model proceeds through multiple stages of inference, incrementally incorporating prior conclusions to guide subsequent reasoning. The architecture integrates frozen visual and textual encoders with an LLM, fine-tuned in a parameter-efficient manner, to generate task-specific answers across diagnostic stages. The overall workflow consists of three phases: multimodal input encoding, context-aware stage-wise reasoning, and autoregressive answer generation.

\begin{figure*}[t]
    \centering
     \includegraphics[width=\textwidth]{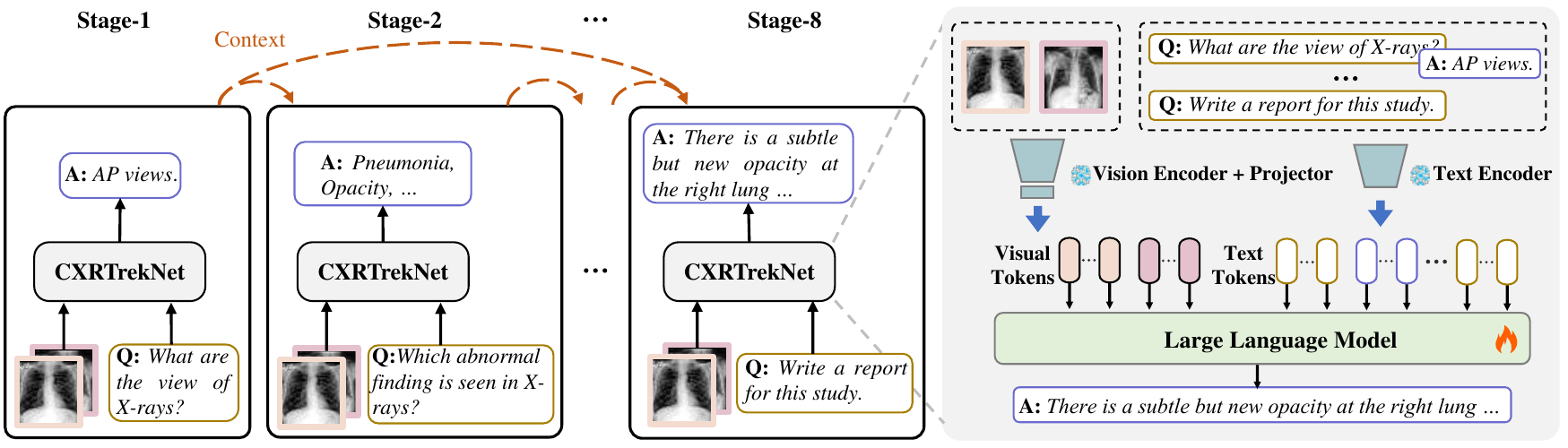}
    \caption{Overview of CXRTrekNet. It takes as input one or more chest X-ray images and a sequence of clinically guided questions. At each stage, it generates an answer by encoding the images through a vision encoder and the Q\&A history through a text encoder. These features are fused by a fine-tuned LLM to generate the current answer. The output is then appended to the context to inform subsequent stages, enabling progressive multi-stage reasoning that mimics radiologist workflows.}
    \label{fig:our_model}
\end{figure*}

\subsection{Multi-modal input encoding}

Each radiological study requiring interpretation for a patient is represented by a set of CXRs $\mathcal{X} = \{ X_1, X_2, \ldots, X_k \}$, which may include both current and previous examinations if any. These images are first processed by a frozen vision encoder $\mathcal{E}_V$ and a visual projection module to produce image-level embeddings. Simultaneously, a sequence of diagnostic questions $\mathcal{Q} = \{ Q_1, Q_2, \ldots, Q_N \}$ is provided, where each $Q_i$ corresponds to a clinically relevant question, such as X-ray quality assessment, abnormality identification, or report generation. 

The textual components, including the current question and previously 
 Q\&As context, are encoded using a frozen text encoder $\mathcal{E}_T$, while images are encoded via the vision encoder $\mathcal{E}_V$. These representations are then composed into an input sequence for the reasoning module.

\subsection{Contextual reasoning across diagnostic stages}

To replicate the cumulative nature of clinical reasoning, CXRTrekNet models the output at each stage $i$ as dependent not only on the current question $Q_i$ and images input $\mathcal{X}$, but also on the historical context accumulated from prior stages. We define this context as:
\begin{equation}
    \texttt{Ctx}_i = \{ Q_1, \hat{\mathcal{A}}_1, \ldots, Q_{i-1}, \hat{\mathcal{A}}_{i-1} \}.
\end{equation}
The model uses a LLM $\mathcal{F}_\phi$, parameterized by $\phi$, to generate the predicted answer $\hat{\mathcal{A}}_i$ at stage $i$. The LLM takes as input the textual and visual features encoded from the current question, the prior context, and all relevant images:
\begin{equation}
    \hat{\mathcal{A}}_i = \mathcal{F}_\phi \left( \mathcal{E}_T(\texttt{Ctx}_i, Q_i); \mathcal{E}_V(X_1, \ldots, X_k) \right),
\end{equation}
where $(\cdot;\cdot)$ refers to the concatenation operation.
This composition ensures that all inputs are ordered chronologically, thereby preserving the stage-wise progression of clinical interpretation.

\subsection{Auto-regressive answer generation and training objective}

For each stage $i$, the model generates the output $\hat{\mathcal{A}}_i$ as a sequence of tokens. This output may correspond to a natural language description, a classification label, a bounding box, or other clinically meaningful responses. We adopt an autoregressive training strategy with Teacher Forcing. Let the ground truth answer be $\mathcal{A}_i = \{ a_{i,1}, a_{i,2}, \ldots, a_{i,|\mathcal{A}_i|} \}$; the stage-wise loss is defined as:
\begin{equation}
    \mathcal{L}_i(\phi) = - \sum_{j=1}^{|\mathcal{A}_i|} \log P(a_{i,j} \mid a_{i,<j}, Q_i, \mathcal{X}, \texttt{Ctx}_i),
\end{equation}
where $a_{i,<j}$ denotes all previously generated tokens\footnote{Notably, for simplicity, we present the stage-wise loss formulation based on a single Q\&A pair per stage. In practice, the number of questions per stage varies across cases.}. The total objective function aggregates losses across all stages:
\begin{equation}
    \mathcal{L}_{\text{overall}}(\phi) = \frac{1}{N} \sum_{i=1}^{N} \mathcal{L}_i(\phi),
\end{equation}
where $N$ is the number of stages.
In practice, we adopt Low-Rank Adaptation (LoRA)~\cite{hu2022lora} for efficient fine-tuning of the LLM $\mathcal{F}_\phi$, while keeping the encoders $\mathcal{E}_T$ and $\mathcal{E}_V$ frozen. This approach significantly reduces computational cost while retaining strong model capacity and domain alignment.

\section{Experiments}
\label{sec:experiment}

\paragraph{Implementation details.}

During training CXRTrekNet, we initialize the visual encoder, text encoder, and LLM with CheXagent~\cite{chen2024chexagent}, and extend RoPE positional encodings~\cite{su2024roformer} to an 8,192-token context window using four-fold linear scaling. To enable parameter-efficient fine-tuning, we apply LoRA~\cite{hu2022lora} with a rank of 256 and an alpha of 512 to all \texttt{torch.linear} layers in LLM except the final token classification head, resulting in approximately 1.5GB of trainable parameters, constituting about 11\% of the full model. Training is conducted about 25k iterations with a batch size of 16 per GPU on four NVIDIA L20 GPUs, totaling around 200 GPU-hours. 

For evaluation, we assess CXRTrekNet on the CXRTrek test set. To further evaluate its generalization across diverse clinical tasks, we additionally evaluate the model on five established benchmarks: Rad-ReStruct~\cite{pellegrini2023rad} and VQA-RAD~\cite{lau2018dataset} for both open-ended and closed-ended VQA, IU-Xray~\cite{demner2016preparing} and MIMIC-CXR~\cite{johnson2019mimic} for radiology report generation, and VinDR-CXR~\cite{nguyen2022vindr} for abnormality classification. More details on the experiments are shown in the Appendix Section A.

\subsection{Comparison with state-of-the-art methods}


\input{assets/Tables/table_CXRTrek_compare_with_sota}

\paragraph{Results on CXRTrek test set.} 

We evaluate CXRTrekNet against five competitive vision-language baselines, including Qwen2-VL~\cite{wang2024qwen2}, XrayGPT~\cite{thawkar2023xraygpt}, MiniGPT-Med~\cite{alkhaldi2024minigpt}, LLaVA-Med v1.5~\cite{li2023llava}, and CheXagent~\cite{chen2024chexagent}, on the CXRTrek test set. As shown in Table~\ref{tab:stagewise_results}, CXRTrekNet achieves the highest overall score, with an average performance of 69.66\%, outperforming the strongest baseline (CheXagent, 33.38\%) by a substantial margin.

Performance improvements are consistently observed in all 8 diagnostic stages. CXRTrekNet demonstrates strong robustness in early-stage tasks, with significant gains in image assessment (Stage-1) and abnormality recognition (Stage-2). In Stage-3, which focuses on fine-grained attribute extraction, it achieves 90.11\% (closed-ended) and 96.26\% (choice) F1 scores—over 25 points higher than all baselines.
In mid-to-late stages, CXRTrekNet shows clear advantages in handling contextual and reasoning-intensive tasks. It obtains a 66.60 BERTScore in Stage-4 (findings’ relationships), outperforming the best baseline by a large margin. In Stage-5 (image comparisons), it also reaches 72.33, indicating a strong temporal reasoning ability. For Stage-6 (future risk questions) and Stage-7 (diagnosis advice), it achieves 69.48 and 53.21 BERTScores, while all baselines remain below 26.
Finally, in Stage-8 (summarizing), CXRTrekNet achieves a BERTScore~\cite{zhang2019bertscore} of 50.52, highlighting its ability to consolidate previous information and generate clinically coherent summaries.

\paragraph{Results on more clinical tasks.}


To further assess the generalizability of CXRTrekNet, we evaluate its performance on five external clinical benchmarks via fine-tuning, covering VQA, abnormality classification, and report generation. Results are summarized in Table~\ref{tab:perf_combined} and Table~\ref{tab:report_generation}.

Specifically, for the VQA task, our model achieves 82.35\% accuracy for Closed-Ended questions and 54.5\% recall for Open-Ended questions on the VQA-RAD benchmark. For the Rad-ReStruct dataset, our model attains an individual Q\&A accuracy of 91.28\% and full report accuracy of 38.60\%.
Secondly, for the abnormality classification task, using the VINDr-CXR dataset, our model achieves 96.06\% classification accuracy and a higher F1 score compared to all baselines.
Thirdly, for the report generation task, the CheXagent model employs a two-stage generation process (generating findings first, then impressions) and a distinct report format. This strategy enhances its clinical evaluation (CE) metrics at the expense of natural language generation (NLG) metrics. Applied to the MIMIC-CXR dataset, while this approach allowed CheXagent to achieve CE metrics slightly surpassing those of our model, our model ultimately demonstrated superior overall performance. On the IU-Xray dataset, our model surpasses all other models in both CE and NLG metrics.
These results demonstrate that training with \textit{CXRTrek} dataset not only boosts performance on primary tasks but also generalizes to diverse radiological interpretation challenges.

\input{assets/Tables/table_vqa_and_classification}

\input{assets/Tables/table_mimic_cxr.tex}

\begin{figure}[t!] 
    \centering 

    \begin{subfigure}[t]{0.48\textwidth}
        \centering 
        \includegraphics[width=\linewidth]{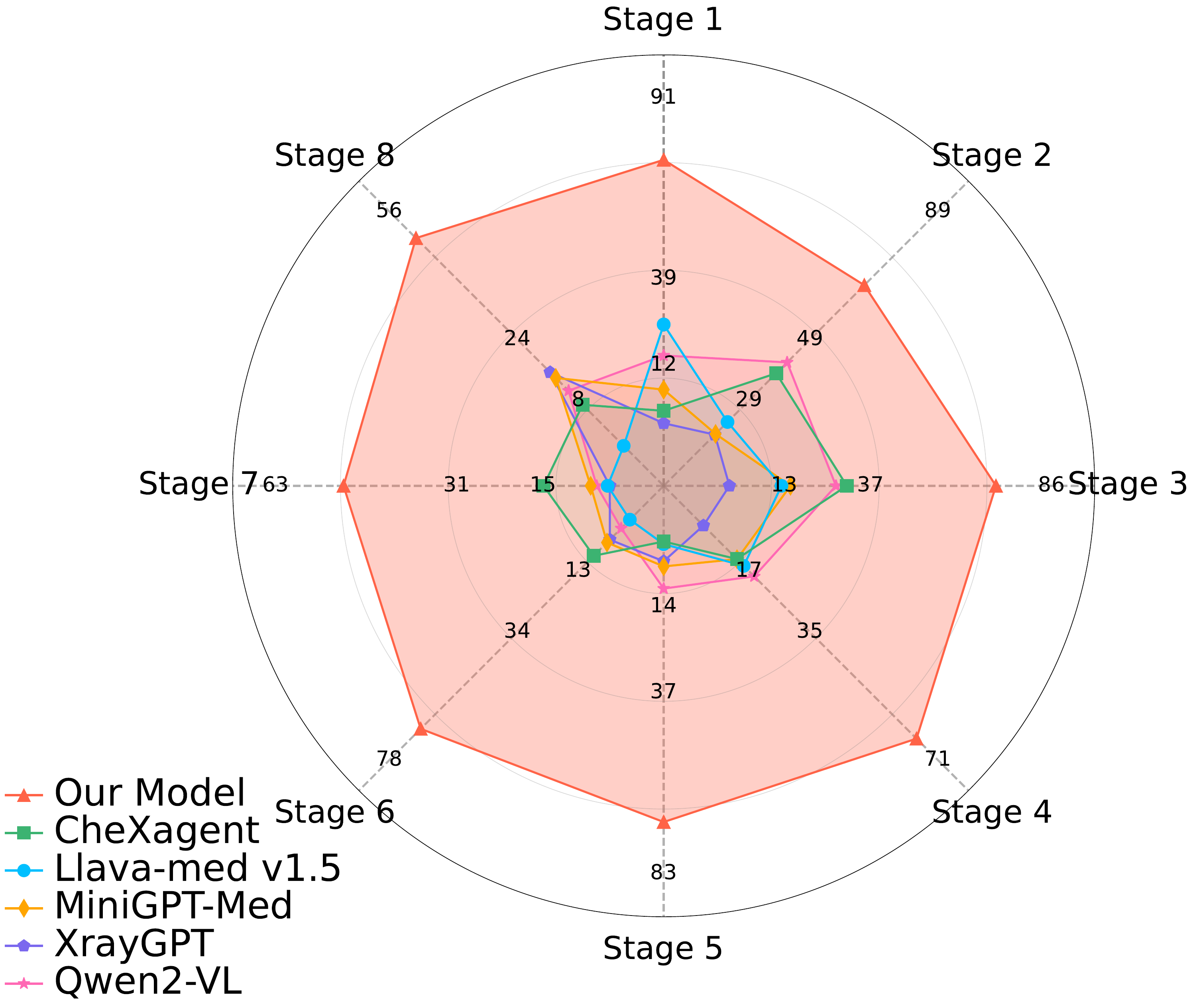}
        \caption{Radar chart comparing the performance of our model with SOTA models across each stage.}
        \label{fig:sota_compare} 
    \end{subfigure}
    \hspace{1pt}
    \begin{subfigure}[t]{0.478\textwidth} 
        \centering 
        \includegraphics[width=\linewidth]{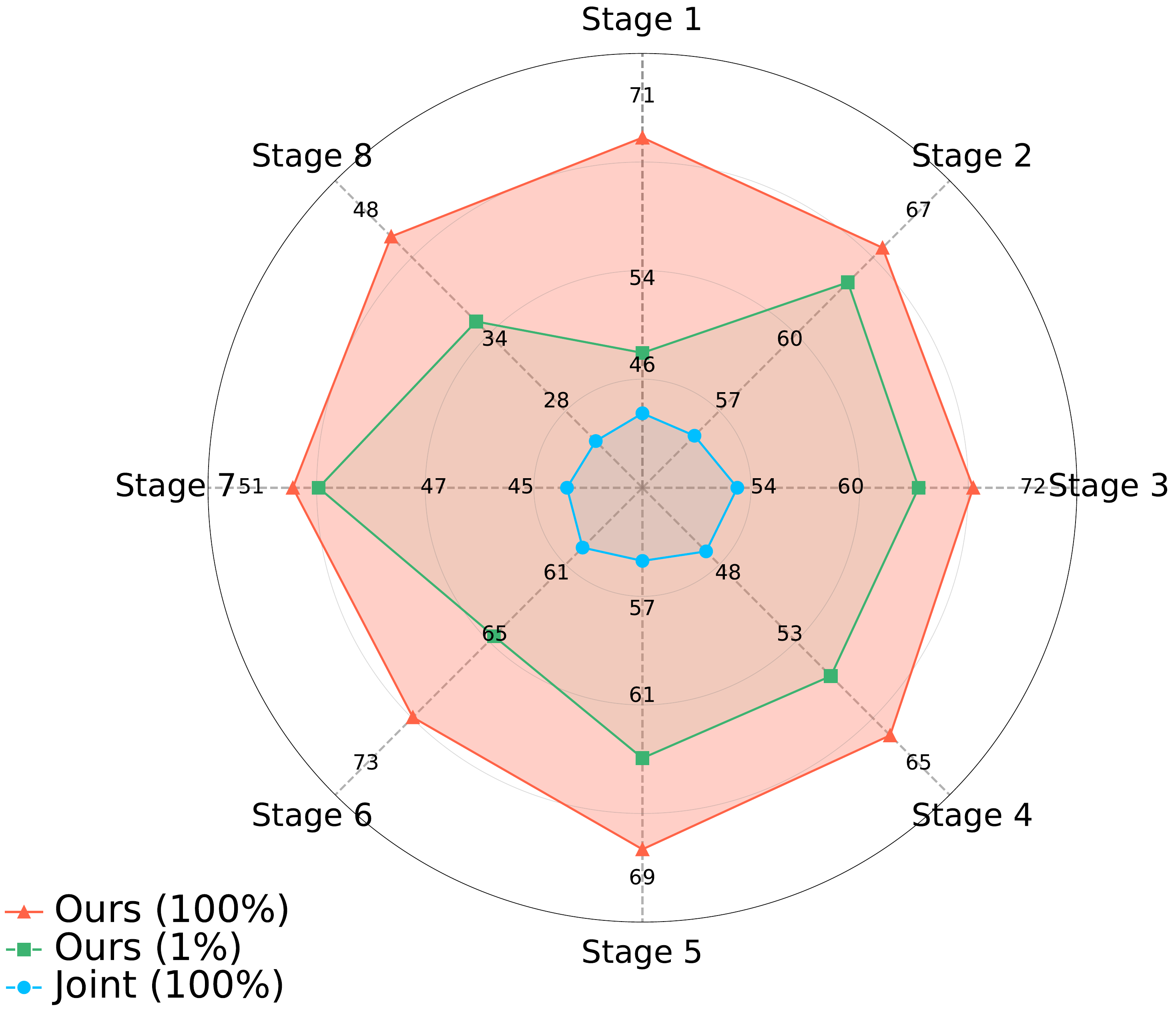}
       \caption{Radar chart comparing model performance under different training configurations in ablation studies.}
        \label{fig:ablation} 
    \end{subfigure}

    \caption{Two radar charts comparing stage-wise performance. Stage scores average corresponding metrics across \textit{Open-Ended}, \textit{Close-Ended}, \textit{Choice}, and \textit{Detection} questions.}
    \label{fig:radar_charts_main} 
\end{figure}

\subsection{Ablation studies}

\paragraph{Analysis of clinical reasoning flow.}
\input{assets/Tables/table_CXRTrek_Ablation}

To further investigate whether the observed performance gains arise from the newly constructed Q\&A pairs or from stage-wise training guided by contextual clinical reasoning, we conduct an ablation study comparing models trained with and without stage-wise training. Specifically, we compare a joint training baseline (Joint), which treats all Q\&A pairs from different stages as independent and trains them in a unified manner, with our CXRTrekNet (Ours), which follows the sequential diagnostic structure defined in CXRTrek.

As shown in Table~\ref{tab:ablation}, although both models are trained on the same amount of data, \textit{Ours (100\%)} consistently outperforms \textit{Joint (100\%)} across all diagnostic stages, achieving a significantly higher average score (64.70 vs. 52.86). Notably, the performance of \textit{Joint (100\%)} even falls below that of \textit{Ours (1\%)}, which is trained on only a fraction of the data. This striking result underscores the effectiveness of stage-wise training guided by the clinical reasoning flow.
The performance gap is especially notable in stages requiring deeper contextual integration. For instance, in Stage-8 (Summarizing), which relies heavily on aggregating previous findings, \textit{Ours (100\%)} achieves a BERTScore of 45.86, significantly surpassing \textit{Joint (100\%)} at 28.86, reflecting a nearly +17\% absolute improvement.



\paragraph{Proportion of training data.}

We further evaluate CXRTrekNet by varying the proportion of training data used—specifically, 1\% and 100\%. The corresponding results are shown in Table~\ref{tab:ablation}. While different configurations may perform best on specific stages, we observe a consistent overall trend: as the amount of training data increases, the model’s average performance across all 8 stages improves steadily. In particular, the average score rises from 60.01 for \textit{Ours (1\%)} to 64.70 for \textit{Ours (100\%)}.
This trend highlights both the high-quality supervision provided by CXRTrek and the benefit of leveraging large-scale data for complex clinical reasoning tasks. A visual comparison of per-stage performance is provided in the radar plot in Figure~\ref{fig:ablation}, further reinforcing the positive correlation between data scale and reasoning ability.

\section{Conclusion}
\label{conclusion}



In this work, we construct CXRTrek, a new large-scale and multi-stage VQA dataset designed to explicitly model the diagnostic reasoning process in clinical CXR interpretation. CXRTrek covers 8 clinically grounded stages and provides fine-grained supervision across over 11 million visual question–answer pairs. Built upon this dataset, we propose CXRTrekNet, a VLLM tailored to follow the sequential logic of clinical workflows. Through extensive experiments on the CXRTrek benchmark and five external datasets, CXRTrekNet demonstrates superior performance across tasks, including classification, detection, VQA, and report generation.
Our results highlight the importance of embedding clinical reasoning flow into medical AI models, leading to better contextual understanding and stronger generalization. 
We will release our dataset and model to facilitate future research in the medical vision-and-language community.


\clearpage
\small  
\bibliographystyle{unsrt}
\bibliography{references}  
\normalsize

\appendix
\clearpage
\setcounter{table}{0}
\setcounter{figure}{0} 
\renewcommand{\thetable}{\Alph{table}}
\renewcommand{\thefigure}{\Alph{figure}}
\renewcommand\theHtable{Appendix.\thetable}
\renewcommand\theHfigure{Appendix.\thefigure}
\newtcolorbox[auto counter]{mybox}[1][]{
  title=My Box \thetcbcounter,
  label=mybox:\thetcbcounter,
  #1
}

\def\mytitle{
Interpreting Chest X-rays Like a Radiologist: \\A Benchmark with Clinical Reasoning
}

\begin{center}
	{
        \Large{\textbf{Supplementary for}}~\Large{\textbf{``\mytitle''}}
	}
\end{center}

\etocdepthtag.toc{mtappendix}
\etocsettagdepth{mtchapter}{none}
\etocsettagdepth{mtappendix}{section}
\etocsettagdepth{mtappendix}{subsection}

{
    \hypersetup{linkcolor=black}
    \footnotesize\tableofcontents
}

\clearpage
\section{More experimental details}
\label{suppl:sec:experiments_details}

This section provides a more detailed account of our experimental procedures, focusing on two primary aspects.
First, we elaborate on the standardized evaluation protocols employed across various datasets and tasks, including specifics on the prompts utilized and the methods for metric computation.
Second, we outline the comprehensive training protocol adopted for our model.

\subsection{Standardized evaluation protocol}


\paragraph{Evaluation of CXRTrek.}

The CXRTrek benchmark employs a pre-defined diagnostic reasoning chain, which establishes a specific sequence of question-answer pairs for evaluation. Due to its multi-stage design, testing is conducted by posing questions sequentially according to this established order. An initial consideration in such evaluations is the direct application of this sequential question-answering paradigm to other existing models (as detailed in Table~\ref{tab:st_mt_cxrtrek_results}). Here, test strategy `Joint' refers to answering questions separately, while `Multi-Stage' involves answering questions with multi-stage contextual information. 
%
Our preliminary investigations indicate that the `Multi-Stage' strategy applied directly to existing models can often degrade their performance, likely due to the models’ limited ability to handle long-range context or stage-dependent reasoning. To ensure fair and meaningful comparison, we report the strongest performance achieved by each baseline model regardless of test strategy, and compare our model against these upper bounds.

\input{assets/Tables/table_st_mt_on_cxrtrek}

For \textit{detection} queries, localization accuracy is measured using the mean Intersection over Union (mIoU). Following methodologies such as those of CheXagent~\cite{smit2020chexbert} and MiniGPT-Med~\cite{alkhaldi2024minigpt}, bounding box coordinates are normalized to a $[0, 100]$ range. For \textit{closed-ended} questions restricted to 'yes'/'no' responses, the TextBlob~\cite{loria2018textblob} library is utilized to determine the semantic polarity (positive or negative) of the answer. In the case of \textit{choice} questions, regular expressions are employed to parse the selected option from the available choices.

\paragraph{Report generation task: MIMIC-CXR and IU-Xray.}


For a fair evaluation of report generation on the MIMIC-CXR~\cite{johnson2019mimic} and IU-Xray~\cite{demner2016preparing} datasets, we employ a prompting strategy consistent with that of most models~\cite{alkhaldi2024minigpt, thawkar2023xraygpt, li2023llava}, wherein one prompt is randomly selected from a designated pool to guide report generation. Specifically, prompts are tailored to target both the "findings" and "impression" sections for MIMIC-CXR, and only the "findings" section for IU-Xray. CheXagent~\cite{smit2020chexbert} is an exception and is evaluated using its native methodology to accommodate its distinct structured output format. Evaluation metrics including BERTScore~\cite{zhang2019bertscore}, Rad-Graph-F1~\cite{jain2021radgraph}, and ROUGE-L~\cite{lin2004rouge} are computed using the CXR-Report-Metrics library~\cite{Yu2022.08.30.22279318}. CheXbert-F1 scores are calculated using the \texttt{f1chexbert} library, which implements the CheXbert methodology~\cite{smit2020chexbert}.

\begin{tcolorbox}[
  title=Prompts for MIMIC-CXR,
  colback=white, 
  colframe=black,
  fonttitle=\bfseries,
  boxrule=0.5pt, 
  arc=2mm,
  left=2mm, 
  right=2mm, 
  top=1mm, 
  bottom=1mm,
  breakable
]
Please provide a report for this X-ray scan as detailed as possible; 
Describe the given chest x-ray image in detail;
Describe the given chest X-ray image in as much detail as possible;
Generate a report for this study based on the X-ray image;
 How would you describe the X-ray image of this patient;   
  Please write the findings and the impression from the X-ray image;
  Please generate a chest X-ray image report as an X-ray export;
  Describe this image in detail;
  Take a look at this image and describe what you notice;
  Please provide a detailed description of the picture.
\end{tcolorbox}

\begin{tcolorbox}[
  title=Prompts for IU-Xray,
  colback=white, 
  colframe=black,
  fonttitle=\bfseries,
  boxrule=0.5pt, 
  arc=2mm,
  left=2mm, 
  right=2mm, 
  top=1mm, 
  bottom=1mm,
  breakable
]
Produce a detailed findings description for the chest X-ray examination as detailed as possible; Generate the findings for the chest X-ray examination as detailed as possible.
\end{tcolorbox}

\paragraph{VQA task: VQA-Rad and Rad-Restruct.}

For our Visual Question Answering (VQA) tasks, evaluation prompts are directly adopted from the VQA-Rad~\cite{lau2018dataset} and Rad-ReStruct~\cite{pellegrini2023rad} datasets. VQA-Rad provides single-turn question-answer pairs, while Rad-ReStruct serves as a multi-turn, structured VQA benchmark. We evaluate models on VQA-Rad following the methodology of~\cite{li2023llava}, and on Rad-ReStruct using its original evaluation settings.

\paragraph{Classification task: VinDr-CXR.}

For the classification task on the VinDr-CXR~\cite{nguyen2022vindr} dataset, we employ a prompting strategy with structured queries to elicit diagnostic information from Vision-Language Models (VLLMs). These prompts encompass direct questions about specific radiological findings (e.g., ``Is there any sign of \{object\} in the X-ray?'', where \{object\} denotes a target condition) and broader inquiries to identify other lesions or diseases beyond a predefined list. Performance is evaluated using Accuracy (Acc) and the Macro F1 score, chosen for their suitability in assessing the categorical outputs (e.g., ``yes''/``no'') generated by VLLMs for such classification questions, as opposed to continuous probability scores.

\begin{tcolorbox}[
  title=Prompts for VinDr-CXR,
  colback=white,  
  colframe=black,
  fonttitle=\bfseries,
  boxrule=0.5pt,  
  arc=2mm,
  left=2mm,  
  right=2mm,  
  top=1mm,  
  bottom=1mm,
  breakable
]
Is there any sign of \{object\} in the X-ray? Is there any sign of other lesions in the X-ray, aside from Aortic enlargement, Atelectasis, Calcification, Cardiomegaly, Clavicle fracture, Consolidation, Edema, Emphysema, Enlarged PA, ILD, Infiltration, Lung Opacity, Lung cavity, Lung cyst, Mediastinal shift, Nodule/Mass, Pleural effusion, Pleural thickening, Pneumothorax, Pulmonary fibrosis, and Rib fracture?  Is there any sign of other diseases in the X-ray, except for COPD, Lung tumor, Pneumonia, and Tuberculosis?
\end{tcolorbox}

\subsection{Detailed  training protocol}

\begin{figure}[h!] 
    \centering
    \includegraphics[width=0.45\linewidth]{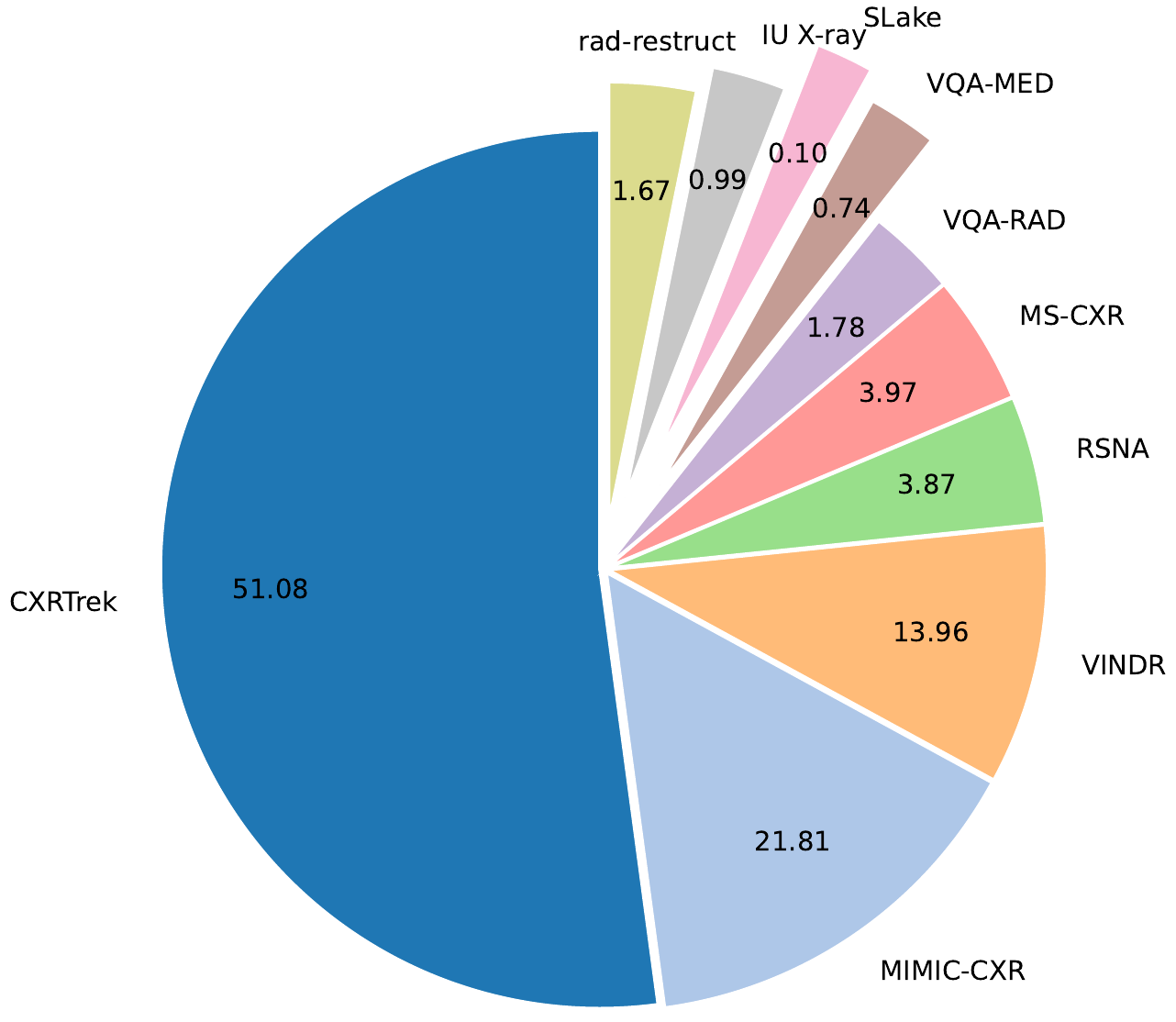} 
    \caption{Distribution of training samples, illustrating the proportional contribution of each dataset to the training mixture.}
    \label{fig:training_data_distribution} 
\end{figure}

For the fine-tuning process, we configure the learning rate to $2 \times 10^{-5}$, utilizing a cosine annealing schedule. The model is initialized from CheXagent, adopting its established conversation template to ensure input consistency. Gradient checkpointing is implemented to optimize memory usage. Training acceleration is achieved through the DeepSpeed~\cite{rasley2020deepspeed} framework in conjunction with bfloat16 (BF16) precision. To mitigate catastrophic forgetting, we adopt a data mixture strategy for training. This approach primarily involved using our proprietary CXRTrek dataset, supplemented with smaller portions of data from other relevant datasets~\cite{rsna-pneumonia-detection-challenge,nguyen2022vindr,johnson2019mimic,demner2016preparing,boecking2022ms,pellegrini2023rad,liu2021slake,ben2019vqa,lau2018dataset}. The mixture proportion of samples from each dataset is visualized in Figure~\ref{fig:training_data_distribution}.

\section{More CXRTrek construction details}
\label{suppl:sec:mtxray_dataset_details}


In this section, we offer a comprehensive elaboration on the construction methodology of the CXRTrek dataset. We begin by presenting the foundational evidence supporting our proposed 8-stage clinical reasoning flow. Subsequently, we detail the diverse range of tasks integrated into each stage of this flow. Finally, we describe the meticulous process involved in designing and constructing each question, including the methodology for annotating the corresponding answers across the different stages.

\subsection{Supporting evidence of our 8-stage clinical reasoning flow}
\label{suppl:subsec:survey_of_guidelines}

Prior to the design of our dataset, we meticulously develop an 8-stage clinical reasoning flow to structure the image interpretation process. This workflow is informed by a comprehensive review of three standard radiology guidelines, supplemented by insights gathered from discussions with practicing clinicians. The primary objective is to ensure that the stages within our dataset closely mirror the authentic radiological reading process. To substantiate the rationale behind each designed stage, we systematically document supporting evidence from these standard guides. This direct alignment is detailed in Table~\ref{tab:table_evidence_from_radpapers}, which presents each of the eight stages alongside corresponding excerpts and justifications derived from these authoritative sources, thereby validating the clinical relevance and sequential logic of our framework.

\input{assets/Tables/table_evidence_from_radpapers}

\subsection{Tasks integrated into CXRTrek dataset}
\label{suppl:subsec:tasks_in_cxrtrek_dataset}

\input{assets/Tables/table_corresponding_tasks}

To develop a comprehensive task taxonomy for our CXRTrek dataset, we survey sixteen public radiology datasets~\cite{irvin2019chexpert,nguyen2022vindr,pellegrini2023rad,ben2019vqa,lau2018dataset,bae2024mimic,liu2021slake,hu2023expert,wu2021chest,bustos2020padchest,national2011national,bannur2023ms,boecking2022ms,humedical,demner2016preparing,rsna-pneumonia-detection-challenge}, cataloging their questions, task types, and clinical objectives. These diverse tasks are consolidated into a unified taxonomy for our Visual Question Answering (VQA) dataset, encompassing all feasible clinical tasks. To ensure annotation precision, question categories like color and counting from datasets such as VQA-RAD~\cite{lau2018dataset} and SLAKE~\cite{liu2021slake} are excluded if their answers are not in the source reports; all other viable task types are incorporated. These tasks are then systematically mapped to eight distinct interpretation stages, defining 10 corresponding task types. An initial set of question templates, generated using GPT-4~\cite{achiam2023gpt} for various scenarios, undergoes rigorous manual review and refinement. This results in 1,040 high-quality templates for question generation within CXRTrek. Further details on the task mapping, interpretation stages, task types, and representative examples are provided in Appendix Table~\ref{tab:table_corresponding_tasks}.

\begin{figure}[t!] 
    \centering
    \includegraphics[width=0.9\linewidth]{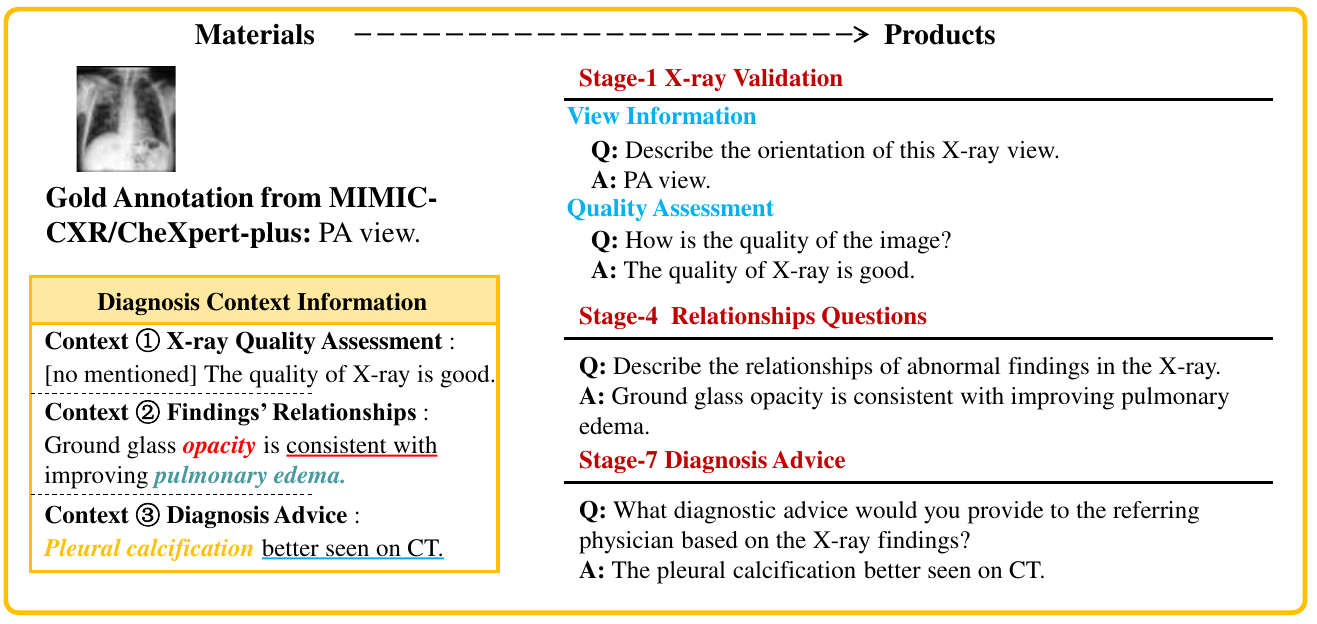}
    \caption{Utilizing annotations from the original datasets (MIMIC-CXR/CheXpert-plus) and diagnostic contextual information extracted via our information extraction framework, the answers on the right-hand side are matched with the corresponding sentences on the left-hand side.}
    \label{fig:stage147_case}
\end{figure}

\begin{figure}[t!] 
    \centering
    \includegraphics[width=0.9\linewidth]{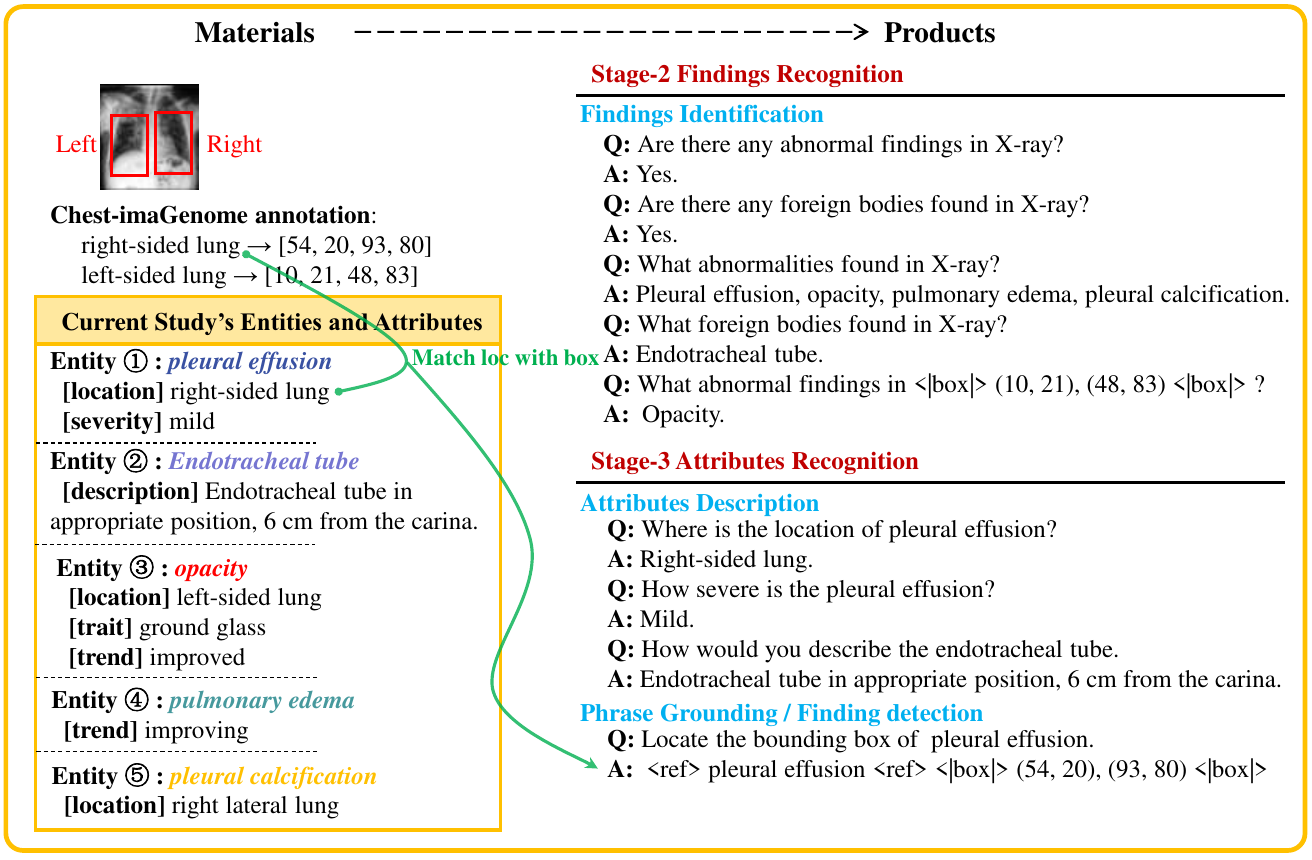}
    \caption{Employing bounding box annotations from Chest-imaGenome, along with entities and attributes extracted through our information extraction framework, the answers on the right-hand side are matched with the corresponding information on the left-hand side.}
    \label{fig:stage23_case}
\end{figure}

\subsection{Annotation of the answer in each stage}
\label{suppl:subsec:more_detais_of_annotation}

\begin{figure}[t!] 
    \centering
    \includegraphics[width=0.9\linewidth]{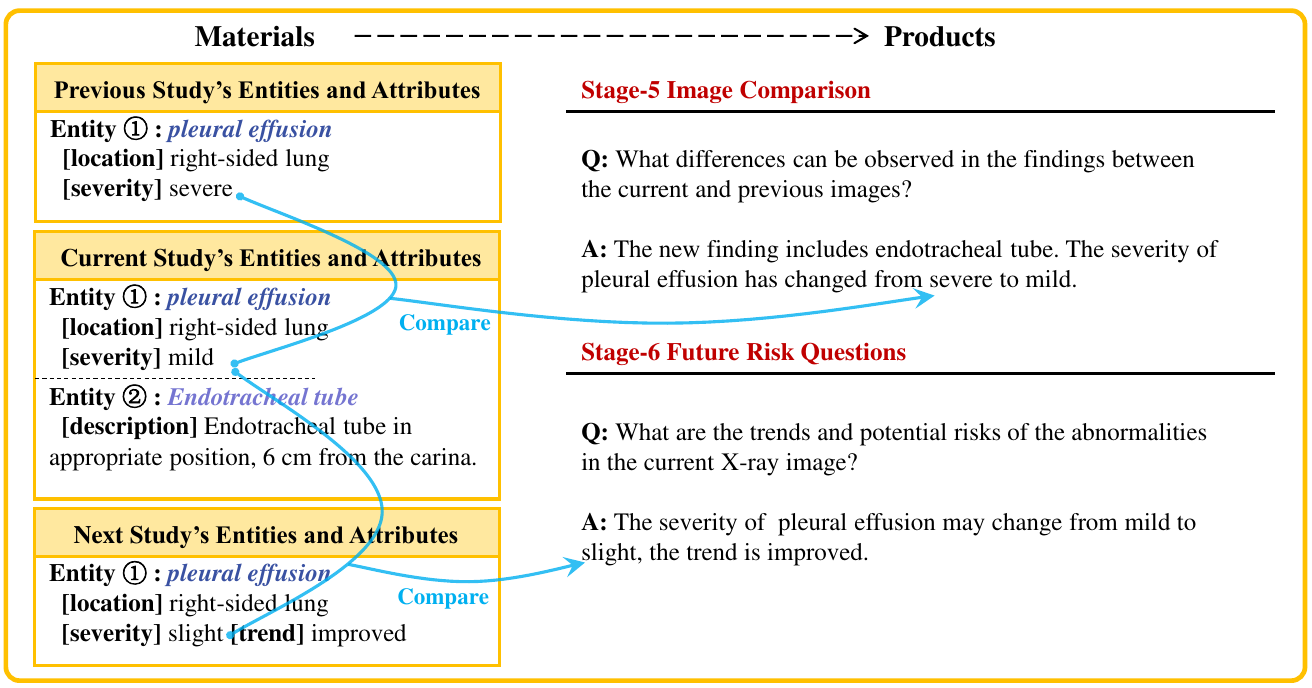}
    \caption{Leveraging entities and attributes extracted by our annotation framework, encompassing information from previous, current, and subsequent studies, the answers on the right-hand side are matched with the corresponding information on the left-hand side.}
    \label{fig:stage56_case}
\end{figure}

This section presents a comprehensive example illustrating the annotation process for answers corresponding to questions at each distinct stage.
Specifically, examples for Stage 1, Stage 4, and Stage 7 are depicted in Figure~\ref{fig:stage147_case}.
Figure~\ref{fig:stage23_case} showcases examples for Stage 2 and Stage 3, while Figure~\ref{fig:stage56_case} presents examples for Stage 5 and Stage 6.
The gold standard for Stage 8 is exclusively derived from the original reports contained in the MIMIC-CXR and CheXpert-plus datasets. To construct the answer for each question, we utilize three primary sources of information: (1) gold standard annotations from MIMIC-CXR~\cite{johnson2019mimic} and CheXpert-plus~\cite{chambon2024chexpert}; (2) bounding box annotations from Chest-imaGenome~\cite{wu2021chest}; and (3) entities, attributes, and diagnostic context extracted from X-ray reports by our dedicated framework. Further details regarding our extraction framework are provided in Appendix Section~\ref{suppl:sec:framework_details}. Subsequently, the answer for each question is systematically constructed from these materials by applying a set of simple logical rules.

\section{The fine-grained information extraction framework}
\label{suppl:sec:framework_details}

\begin{figure}[t!]
    \centering
    \includegraphics[width=\linewidth]{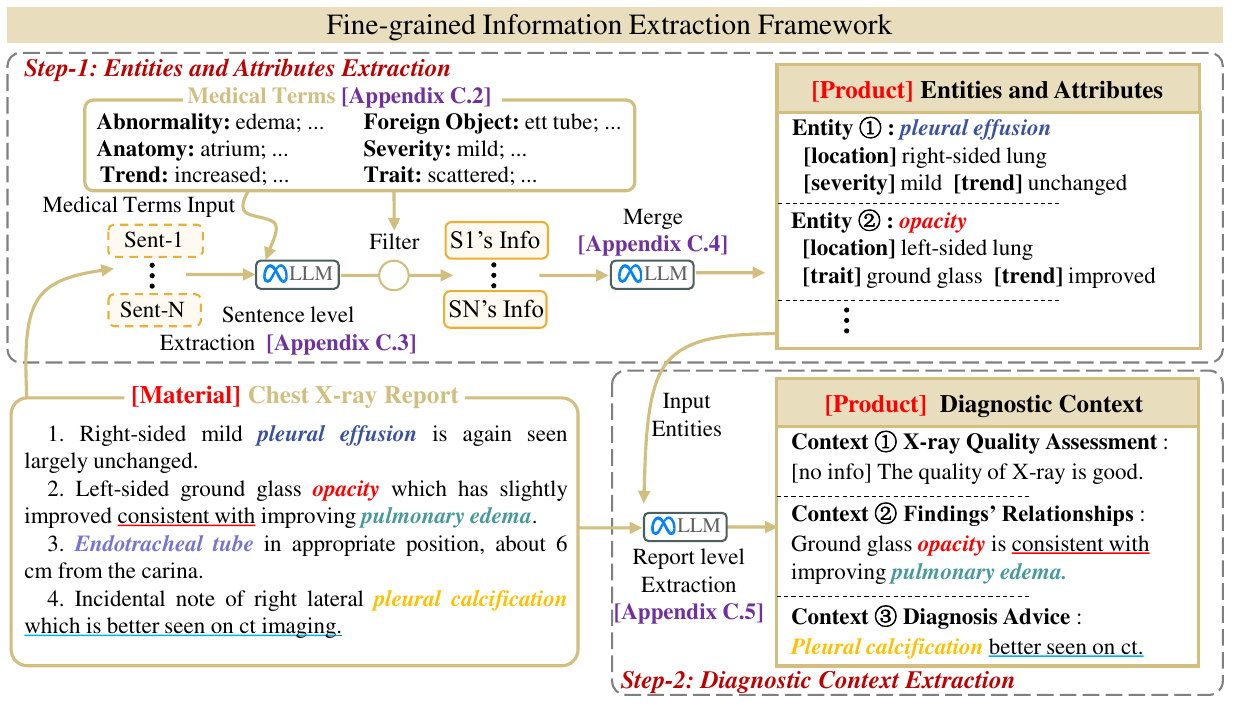}
    \caption{The fine-grained information extraction framework, which extracts entities, attributes, and three distinct diagnostic contextual information from reports using LLM and prompt engineering.}
    \label{fig:information_extract_framework}
\end{figure}

In this section, we elaborate on the design and implementation of our fine-grained information extraction framework. We systematically describe the key procedures depicted in Figure~\ref{fig:information_extract_framework}: \texttt{Sentence-Level Extraction}, \texttt{Report-Level Extraction}, the \texttt{Merge} process, and the \texttt{Medical Terms} vocabulary. This approach ensures both methodological rigor and domain validity. Applying this framework, we annotate reports from the MIMIC-CXR~\cite{bae2024mimic} and CheXpert-plus~\cite{chambon2024chexpert} datasets, extracting a total of 415,326 items. These items are structured using a custom dictionary format, detailed in Appendix~\ref{suppl:subsec:extracted_dictionary}.

To validate the quality of the automatically extracted labels from our framework, we conduct a human comparison. We randomly select 100 samples from the training set. For each sample, human annotators manually extract all labels targeted by the model. We then compare the model-extracted outputs with these human references, achieving a high BERTScore of 92.53. This result validates the reliability of our annotation extraction framework. Detailed validation for each field of the dictionary is presented in Table~\ref{tab:check_extract}.

\input{assets/Tables/table_check_extract}

\subsection{Extracted information dictionary}
\label{suppl:subsec:extracted_dictionary}

For each radiology report, our framework outputs a structured data dictionary in JSON format, composed of five primary components: \texttt{abnormalities}, \texttt{foreign\_bodies}, \texttt{relationships}, \texttt{quality\_issue}, and \texttt{recommendation}. To facilitate readability, we display the schema as a list, annotating each element with its type and a brief description of its contents. The complete specification is shown below:

\begin{tcolorbox}[
  title=Extracted Fine-grained Information Dictionary,
  colback=white, colframe=black,
  fonttitle=\bfseries,
  boxrule=0.5pt, arc=2mm,
  left=2mm, right=2mm, top=1mm, bottom=1mm,
  breakable
]
\begin{description}
  \item[\textbf{abnormalities}]
    \begin{itemize}
      \item[\textbf{:}]  \texttt{List[Dict]}
      \item \textbf{Name:} \texttt{String}
      \item \textbf{Pre\_or\_Absent:} \texttt{String} [Note: The present status of the entity]
      \item \textbf{Location:} \texttt{List[String]}
      \item \textbf{Severity:} \texttt{List[String]}
      \item \textbf{Trend:} \texttt{List[String]}
      \item \textbf{Trait:} \texttt{List[String]}
      \item \textbf{Rel\_to\_Other:} \texttt{List[String]} [Note: Relationships with other entities]
      \item \textbf{Description:} \texttt{List[String]}
    \end{itemize}
  \item[\textbf{foreign\_bodies}]
    \begin{itemize}
      \item[\textbf{:}]  \texttt{List[Dict]}
      \item \textbf{Name:} \texttt{String}
      \item \textbf{Pre\_or\_Absent:} \texttt{String} 
      \item \textbf{Location:} \texttt{List[String]}
      \item \textbf{Trait:} \texttt{List[String]}
      \item \textbf{Rel\_to\_Other:} \texttt{List[String]} 
      \item \textbf{Description:} \texttt{List[String]}
    \end{itemize}
  \item[\textbf{relationships}]
    \begin{itemize}
      \item[\textbf{:}]  \texttt{List[String]}
      \item Sentence(s) describing relationships among entities
    \end{itemize}
  \item[\textbf{quality\_issue}]
    \begin{itemize}
      \item[\textbf{:}]  \texttt{List[String]}
      \item Sentence(s) describing any X‑ray quality issues
    \end{itemize}
  \item[\textbf{recommendation}]
    \begin{itemize}
      \item[\textbf{:}]  \texttt{List[String]}
      \item Sentence(s) providing recommended follow‑up or actions
    \end{itemize}
\end{description}
\end{tcolorbox}

\subsection{Medical Terms}
\label{suppl:subsec:medical_terms}

To mitigate the risk of hallucination and to ensure clinical validity in the outputs of our information annotation framework, we introduce a Medical Knowledge Guarantee module that relies on two complementary resources: a curated medical terminology lexicon and a comprehensive synonym dictionary. By anchoring extracted entities to these resources, we both constrain the model’s outputs to known medical concepts and prevent redundant or conflicting labels.

\paragraph{Medical terminology.}  

We construct a specialized medical vocabulary by integrating the UMLS Metathesaurus~\cite{bodenreider2004unified}, which contains over three million unique concept names across more than 200 semantic types, with manually curated term lists derived from public radiology corpora.
To identify terms present within our dataset, a trie-based string-matching algorithm is applied to a corpus of over 600{,}000 de-identified chest X-ray reports from MIMIC-CXR~\cite{johnson2019mimic} and CheXpert-plus~\cite{chambon2024chexpert}. Only terms occurring at least ten times are retained for inclusion.
These high-frequency terms are then merged with our manually compiled lexicon, resulting in a final vocabulary of 5,030 terms.
Subsequently, this vocabulary is used in our prompt engineering strategy, specifically within the \textit{Sentence-level Extraction} in the extraction framework and \textit{Filter} components in the postprocess. These components utilize the vocabulary to identify and filter entities that demonstrate significant token overlap with the terms in our established vocabulary of medical terms.

Each term is assigned to one of nine semantic categories: \texttt{relation\_terms}, \texttt{abnormalities}, \texttt{foreign\_bodies}, \texttt{anatomy}, \texttt{direction}, \texttt{trend}, \texttt{severity}, \texttt{trait}, and \texttt{other}.

\begin{tcolorbox}[
  title=Nine category terms,
  colback=white, colframe=black,
  fonttitle=\bfseries,
  boxrule=0.5pt, arc=2mm,
  left=2mm, right=2mm, top=1mm, bottom=1mm,
  breakable
]
\begin{description}[leftmargin=1.5cm]
  \item[\textbf{relation\_terms}]  
  Expressions that connect findings with anatomical sites or other findings ($e.g.$, “associated with”, “concerned with”). They help determine the structure of extracted relationships.

  \item[\textbf{abnormalities}]  
  Descriptions of pathological conditions ($e.g.$, “effusion”, “opacity”, “collapse”), which are the primary targets for information extraction.

  \item[\textbf{foreign\_bodies}]  
  Mentions of medical devices or exogenous objects visible in the image ($e.g.$, “central line”, “pacemaker”). Recognizing these helps avoid false positive findings.

  \item[\textbf{anatomy}]  
  References to anatomical structures ($e.g.$, “lung”, “diaphragm”, “mediastinum”) that serve as anchors for locating findings.

  \item[\textbf{direction}]  
  Terms indicating spatial orientation or position ($e.g.$, “left”, “right”, “upper”, “lower”), used to qualify anatomical location.

  \item[\textbf{trend}]  
  Words that describe temporal progression or change ($e.g.$, “improved”, “worsening”, “unchanged”) and support longitudinal reasoning.

  \item[\textbf{severity}]  
  Terms denoting the degree or extent of findings ($e.g.$, “mild”, “moderate”, “severe”), enabling quantification of clinical impact.

  \item[\textbf{trait}]  
  Descriptive adjectives that characterize visual or morphological features ($e.g.$, “spiculated”, “calcified”, “linear”) and enhance specificity.

  \item[\textbf{other}]  
  A miscellaneous category for terms that do not clearly fall into the other classes but are still informative for parsing and interpretation.
\end{description}
\end{tcolorbox}

This structured categorization ensures consistent labeling throughout the framework and enables downstream modules to interpret each term according to its clinical role, thereby improving both extraction precision and interpretability.

\paragraph{Meidcal synonym judge.}

To avoid information redundancy and to ensure that semantically equivalent expressions are treated uniformly, we construct a synonym dictionary using two strategies. First, we extract all pairs of terms sharing the same UMLS Concept Unique Identifier (CUI), yielding 11,648 high‑confidence synonym pairs ($e.g.$, ``heart enlarged'' and ``cardiomegaly''). Second, to capture variations not covered by UMLS, we apply a string‑similarity filter based on Python’s \texttt{difflib}\footnote{https://docs.python.org/3/library/difflib.html} library, designating any pair with a similarity score above 0.90 as synonyms ($e.g.$, ``mediastinal widening'' and ``mediastinum widened'', ``rib fracture'' and ``fractured rib''). During both the sentence‑level merging of extracted attributes and the generation of question–answer pairs, we consult this dictionary to collapse synonymous mentions. For example, without synonym collapsing, a report noting ``heart enlarged'' could lead to a Q\&A pair with the question ``Are there any cardiomegaly findings?'' and an incorrect answer ``No''. By mapping ``heart enlarged'' to ``cardiomegaly'', our method correctly yields ``Yes'', thus avoiding such contradictory annotations.

\subsection{Sentence-Level Extraction}

In this phase, we extract entities representing \textbf{abnormality} and \textbf{foreign objects} from each sentence, along with their associated attributes: \textbf{location}, \textbf{trait}, \textbf{trend}, \textbf{severity}, and \textbf{description}. To determine whether each entity is semantically present or absent, we introduce an additional \textbf{present\_or\_absent} field. The pure-text Llama 3.2 8B large language model~\cite{grattafiori2024llama} is employed for information extraction. Through empirical analysis of prompt engineering, we observe that entity extraction at the sentence level achieves higher precision compared to other approaches. Specifically, we design a series of structured prompts to guide the large language model (LLM) in extracting the corresponding information. Several representative prompt designs are presented below.

\paragraph{Prompt for extracting entities.}

We employ the following prompt to extract information regarding abnormalities and foreign objects:

\begin{tcolorbox}[
  title=Nine Category Terms,
  colback=white, 
  colframe=black,
  fonttitle=\bfseries,
  boxrule=0.5pt, 
  arc=2mm,
  left=2mm, 
  right=2mm, 
  top=1mm, 
  bottom=1mm,
  breakable
]
Your task is to extract the abnormality and foreign body entities from the provided X-ray report.\\
- Report: ``\{0\}''

Here are the initial entities for reference, but prioritize the definitions and rules below:\\
- Initial Abnormalities: \{1\}\\
- Initial Foreign Bodies: \{2\}

\textbf{Definitions:}
- \textbf{Abnormalities}: Specific medical conditions, diseases, injuries, abnormal shapes, opacities and increased densities, or signs of abnormality in the X-ray, excluding normal anatomical terms.
- \textbf{Foreign Bodies}: Objects or markers not naturally part of the body, such as medical devices or post-operation markers.

\textbf{Extraction Rules:}
\begin{enumerate}
    \item \textbf{Match}: Entities must match the abnormalities and foreign bodies described in the X-ray report.
    \item \textbf{Exclude Descriptors and Anatomy}: Do not include size, degree, location, numerical measurements, or anatomy unless they are part of a specific abnormality term.
    \item \textbf{No Inference}: Only include entities explicitly mentioned in the report.
\end{enumerate}

First, output the analysis in the following format:\\
\textbf{Analysis}: [Output analysis of all possible abnormalities and foreign bodies descriptions based on the definitions and rules; if not found, state that clearly in short.]

Second, output all valid entities in the following format, using semicolons (;) to separate entities, and output ``None'' if no valid entities are found:\\
\textbf{Abnormalities}: [entity\_1]; [entity\_2]; \dots \\
\textbf{Foreign Bodies}: [entity\_1]; [entity\_2]; \dots

Stop immediately after listing the entities.
\end{tcolorbox}

\paragraph{Prompt for presence and absence status classification.}

We employ the following prompt to classify the presence and absence status of an entity:

\begin{tcolorbox}[
  title=Present and Absent Status Classification,
  colback=white, 
  colframe=black,
  fonttitle=\bfseries,
  boxrule=0.5pt, 
  arc=2mm,
  left=2mm, 
  right=2mm, 
  top=1mm, 
  bottom=1mm,
  breakable
]
Based on the chest X-ray report: ``\{0\}'', do you think that ``\{1\}'' is present or absent in the patient's body?

\textbf{Definitions:}
\begin{itemize}
    \item \textbf{Present:}
    \begin{itemize}
        \item \textbf{Possible present}: Terms such as positive possible, likely, represents, consistent, indicative, or compatible with in relation to the entity.
        \item \textbf{Absolutely present}: 
        \begin{itemize}
            \item Terms such as present, no change, remains unchanged, inserted, or placed in relation to the entity.
            \item If an entity is described in a declarative sentence without any related terms, it is considered Absolutely Present by default.
        \end{itemize}
    \end{itemize}
    \item \textbf{Absent:}
    \begin{itemize}
        \item \textbf{Possible absent}: Terms such as less likely, negative possible, or not likely in relation to the entity.
        \item \textbf{Absolutely absent}: Terms such as absent, removed, cleared, denied, excluded, healed, resolved, ruled out, or not present in relation to the entity.
    \end{itemize}
\end{itemize}

First, output the analysis based on the definitions in the following format:\\
\textbf{Analysis}: [Output the analysis focusing on the ``\{1\}'' entity with evidence from definitions clearly and concisely.]

Second, output is limited to a word, present or absent.\\
\textbf{The status of ``\{1\}'':} [present or absent]

Stop immediately after outputting the results.
\end{tcolorbox}

\paragraph{Prompts for location attributes extraction.}

We employ the following prompt to extract the location attributes of an entity:

\begin{tcolorbox}[
  title=Location Attributes Extraction Output,
  colback=white, 
  colframe=black,
  fonttitle=\bfseries,
  boxrule=0.5pt, 
  arc=2mm,
  left=2mm, 
  right=2mm, 
  top=1mm, 
  bottom=1mm,
  breakable
]
Please answer the following question as an X-ray assistant without referencing the report or mentioning itself:\\
\textbf{Question}: ``What is the detailed location of \{0\}?''\\
\textbf{Answer}: One sentence or None.

Check the chest X-ray report: ``\{1\}''\\
Here are some location keywords for reference: ``\{2\}'' but prioritize the rules below.

\textbf{Rules:}
\begin{enumerate}
    \item Location is not part of the entity name (``\{0\}'').
    \item Do not infer an answer that is not explicitly mentioned in the report.
    \item Location must be detailed and complete with anatomy, location, and other direction and distance information if possible.
\end{enumerate}

\textbf{Location}: None
\end{tcolorbox}

\paragraph{Prompt for severity attributes extraction.}

We employ the following prompt to extract the severity attributes of an entity:

\begin{tcolorbox}[
  title=Severity Attributes Extraction,
  colback=white, 
  colframe=black,
  fonttitle=\bfseries,
  boxrule=0.5pt, 
  arc=2mm,
  left=2mm, 
  right=2mm, 
  top=1mm, 
  bottom=1mm,
  breakable
]
Your task is to extract the severity information for entities from the chest X-ray report and output the severity based on the provided severity list.

The chest X-ray report: ``\{0\}''\\
The entity list: \{1\}\\
The severity list is as follows: \{2\}

\textbf{Rules:}
\begin{enumerate}
    \item Do not infer severity information from the report if it is not explicitly mentioned.
    \item The severity must be from the provided severity list.
    \item The severity must directly link to the entity in the report.
    \item Connection words such as ``and'', ``or'' can transfer the severity information.
\end{enumerate}

First, you need to analyze the severity information for each entity:\\
\textbf{Analysis}: [Output the analysis of the severity information for each entity in one sentence, strictly using the report's language. If no severity is mentioned, clearly state this.]

Second, you need to output the severity information for each entity in the following format, using semicolons (;) to separate entities and a vertical line (|) to separate entities and severity information, leave it empty if no valid entities are found:\\
\textbf{Severity}: [entity1|severity1; entity2|severity2; \dots]

Stop immediately after outputting the result.
\end{tcolorbox}

\paragraph{Prompt for trend attributes extraction.}

We employ the following prompt to extract the trend attributes of an entity:

\begin{tcolorbox}[
  title=Trend Attributes Extraction,
  colback=white, 
  colframe=black,
  fonttitle=\bfseries,
  boxrule=0.5pt, 
  arc=2mm,
  left=2mm, 
  right=2mm, 
  top=1mm, 
  bottom=1mm,
  breakable
]
Your task is to extract the trend information for entities from the chest X-ray report and output the trend based on the provided trend list.

The chest X-ray report: ``\{0\}''\\
The entity list: \{1\}\\
The trend list is as follows: \{2\}

\textbf{Rules:}
\begin{enumerate}
    \item Do not infer trend information from the report if it is not explicitly mentioned.
    \item The trend must be from the provided trend list.
    \item The trend must directly link to the entity in the report.
    \item Connection words such as ``and'', ``or'' can transfer the trend information.
\end{enumerate}

First, you need to analyze the trend information for each entity:\\
\textbf{Analysis}: [Output the analysis of the trend information for each entity in one sentence, strictly using the report's language. If no trend is mentioned, clearly state this.]

Second, you need to output the trend information for each entity in the following format, using semicolons (;) to separate entities and a vertical line (|) to separate entities and trend information, leave it empty if no valid entities are found:\\
\textbf{Trend}: [entity1|trend1; entity2|trend2; \dots]

Stop immediately after outputting the result.
\end{tcolorbox}

\paragraph{Prompt for trait attributes extraction.}

We employ the following prompt to extract the trait attributes of an entity:

\begin{tcolorbox}[
  title=Trait Attributes Extraction,
  colback=white, 
  colframe=black,
  fonttitle=\bfseries,
  boxrule=0.5pt, 
  arc=2mm,
  left=2mm, 
  right=2mm, 
  top=1mm, 
  bottom=1mm,
  breakable
]
Your task is to extract the trait information for entities from the chest X-ray report and output the trait based on the provided trait list.

The chest X-ray report: ``\{0\}''\\
The entity list: \{1\}\\
The trait list is as follows: \{2\}

\textbf{Rules:}
\begin{enumerate}
    \item Do not infer trait information from the report if it is not explicitly mentioned.
    \item The trait must be from the provided trait list.
    \item The trait must directly link to the entity in the report.
    \item Connection words such as ``and'', ``or'' can transfer the trait information.
\end{enumerate}

First, you need to analyze the trait information for each entity:\\
\textbf{Analysis}: [Output the analysis of the trait information for each entity in one sentence, strictly using the report's language. If no trait is mentioned, clearly state this.]

Second, you need to output the trait information for each entity in the following format, using semicolons (;) to separate entities and a vertical line (|) to separate entities and trait information, leave it empty if no valid entities are found:\\
\textbf{Trait}: [entity1|trait1; entity2|trait2; \dots]

Stop immediately after outputting the result.
\end{tcolorbox}

\paragraph{Prompt for description attributes extraction.}

We employ the following prompt to extract the description attributes of an entity:

\begin{tcolorbox}[
  title=Description Attributes Extraction,
  colback=white, 
  colframe=black,
  fonttitle=\bfseries,
  boxrule=0.5pt, 
  arc=2mm,
  left=2mm, 
  right=2mm, 
  top=1mm, 
  bottom=1mm,
  breakable
]
Please answer the following question based on the report without referencing the report or mentioning itself:\\
\textbf{Question}: ``What is the detailed description of \{0\}?''\\
\textbf{Answer}: One sentence or None.

The chest X-ray report: ``\{1\}''

\textbf{Rules:}
\begin{enumerate}
    \item Only focus on \{0\}'s description, exclude the information of other abnormalities and foreign bodies.
    \item Do not infer the information from the report if it is not explicitly mentioned.
    \item The description must be detailed and complete, including the shape, size, density, and other characteristics if possible.
\end{enumerate}

\textbf{Description}: [Answer one sentence about the detailed and complete description of \{0\} using the report's language]

Stop immediately after outputting the results without any additional information.
\end{tcolorbox}

\subsection{Sentence-Level Information Merge}

Within a single report, multiple sentences may describe identical or differing attributes of the same entity. For example, an initial sentence might state, ``We found a severe pleural effusion in the left hemithorax'', yielding the entity ``pleural effusion'', its location as ``left hemithorax'', and severity as ``severe''. A subsequent sentence could then read, ``The pleural effusion on the right side is slight'', identifying the same entity ``pleural effusion'', but with location ``right hemithorax'' and severity ``slight''. Our process then merges these attribute mentions: the consolidated entity becomes ``pleural effusion'' with location ``bilateral hemithoraces'' and severity ``severe on the left, slight on the right''. In essence, this step consolidates the precise but scattered information extracted at the sentence level. We now present the series of structured prompts employed in this phase.

To implement the \textit{Merge} process for entities and attributes, we adopt a two-stage approach. First, we utilize pre-built synonym pairs (detailed in Appendix~\ref{suppl:subsec:medical_terms}) in conjunction with string-matching algorithms specifically to merge entities and attributes that exhibit synonymous characteristics. Subsequently, for more complex scenarios not fully resolved by this initial stage, we leverage Large Language Models (LLMs) and prompt engineering to achieve a comprehensive merge of the entities and attributes.

\paragraph{Prompt for merge location attributes.}

We employ the following prompt to refine and polish the location attributes of an entity:

\begin{tcolorbox}[
  title=Merge Location Attributes,
  colback=white, 
  colframe=black,
  fonttitle=\bfseries,
  boxrule=0.5pt, 
  arc=2mm,
  left=2mm, 
  right=2mm, 
  top=1mm, 
  bottom=1mm,
  breakable
]
You are a medical AI assistant. Your task is to refine and polish the location information of \{0\}.

The focus target is: ``\{0\}''.\\
The extracted location information is: ``\{1\}''.\\
The chest X-ray radiological fact states: ``\{2\}''.

\textbf{Rules:}
\begin{enumerate}
    \item Correct the error or conflict information.
    \item Polish the unspecific location.
    \item If there are no valid and correct location specified, state that clearly.
    \item Focus on the location of \{0\} only.
    \item If the location can be merged, please summarize them into one word or a sentence, such as left and right should be merged.
\end{enumerate}

\textbf{Output in the following format with <answer> and <think> tags:}\\
\texttt{<think>}\\
Thinking process to analyze and refine the content in one sentence; If no valid and correct location specified, clearly state this.\\
\texttt{</think>}\\
\texttt{<answer>}\\
Output the detailed location-only information using the report's language from the view of diagnosis without mentioning the fact itself in one sentence or few words: [Location]; If no valid and correct location specified, output ``None''.\\
\texttt{</answer>}

Stop immediately after outputting the results.
\end{tcolorbox}

\paragraph{Prompt for merge traits attributes.}

We employ the following prompt to refine and polish the traits attributes of an entity:

\begin{tcolorbox}[
  title=Merge Traits Attributes,
  colback=white, 
  colframe=black,
  fonttitle=\bfseries,
  boxrule=0.5pt, 
  arc=2mm,
  left=2mm, 
  right=2mm, 
  top=1mm, 
  bottom=1mm,
  breakable
]
You are a medical AI assistant. Your task is to refine and polish the traits information of \{0\}.\\
Traits are the characteristics/features/attributes of the \{0\}.

The focus target is: ``\{0\}''.\\
The extracted traits information is: ``\{1\}''.\\
The chest X-ray radiological fact states: ``\{2\}''.

\textbf{Rules:}
\begin{enumerate}
    \item Correct the error or conflicting traits to find the suitable traits.
    \item Polish the unspecific traits with more specific located information.
    \item If there are no valid traits specified, state that clearly.
    \item Focus on the traits of \{0\} only.
    \item Merge the traits with the same location information into a combined word.
\end{enumerate}

\textbf{Output in the following format with <answer> and <think> tags:}\\
\texttt{<think>}\\
Thinking process to analyze and refine the content in one concise and short sentence; If no valid and correct traits specified, clearly state this.\\
\texttt{</think>}\\
\texttt{<answer>}\\
Output the traits information separated by semicolon (;) from the view of diagnosis without mentioning the fact itself [Trait-1 with specific information-1; Trait-2 with specific information-2; \dots]; If no valid and correct traits specified, output ``None''.\\
\texttt{</answer>}

Stop immediately after outputting the results.
\end{tcolorbox}

\paragraph{Prompt for merge severity attributes.}

We employ the following prompt to refine and polish the severity attributes of an entity:

\begin{tcolorbox}[
  title=Merge Severity Attributes,
  colback=white, 
  colframe=black,
  fonttitle=\bfseries,
  boxrule=0.5pt, 
  arc=2mm,
  left=2mm, 
  right=2mm, 
  top=1mm, 
  bottom=1mm,
  breakable
]
You are a medical AI assistant. Your task is to refine and polish the severity information of \{0\}.

The focus target is: ``\{0\}''.\\
The extracted severity information is: ``\{1\}''.\\
The chest X-ray radiological fact states: ``\{2\}''.

\textbf{Rules:}
\begin{enumerate}
    \item Correct the error or conflicting severity to find the suitable severity.
    \item Polish the unspecific severity with more specific located information.
    \item Focus on the severity of \{0\} only.
    \item Merge the severity with the same location information into a combined word.
\end{enumerate}

\textbf{Output in the following format with <answer> and <think> tags:}\\
\texttt{<think>}\\
Thinking process to analyze and refine the content in one concise and short sentence; If no valid and correct severity specified, clearly state this.\\
\texttt{</think>}\\
\texttt{<answer>}\\
Output the severity information separated by semicolon (;) from the view of diagnosis without mentioning the fact itself [Severity-1 with specific information-1; Severity-2 with specific information-2; \dots]; If no valid and correct severity specified, output ``None''.\\
\texttt{</answer>}

Stop immediately after outputting the results.
\end{tcolorbox}

\paragraph{Prompt for merge trend attributes.}

We employ the following prompt to refine and polish the trend attributes of an entity:

\begin{tcolorbox}[
  title=Merge Trend Attributes,
  colback=white, 
  colframe=black,
  fonttitle=\bfseries,
  boxrule=0.5pt, 
  arc=2mm,
  left=2mm, 
  right=2mm, 
  top=1mm, 
  bottom=1mm,
  breakable
]
You are a medical AI assistant. Your task is to refine and polish the trend information of \{0\}.

The focus target is: ``\{0\}''.\\
The extracted trend information is: ``\{1\}''.\\
The chest X-ray radiological fact states: ``\{2\}''.

\textbf{Rules:}
\begin{enumerate}
    \item Correct the error or conflicting trend to find the suitable trend.
    \item Polish the unspecific trend with more specific located information.
    \item Focus on the trend of \{0\} only.
    \item Merge the trend with the same location information into a combined word.
\end{enumerate}

\textbf{Output in the following format with <answer> and <think> tags:}\\
\texttt{<think>}\\
Thinking process to analyze and refine the content in one concise and short sentence; If no valid and correct trend specified, clearly state this.\\
\texttt{</think>}\\
\texttt{<answer>}\\
Output the trend information separated by semicolon (;) from the view of diagnosis without mentioning the fact itself [Trend-1 with specific information-1; Trend-2 with specific information-2; \dots]; If no valid and correct trend specified, output ``None''.\\
\texttt{</answer>}

Stop immediately after outputting the results.
\end{tcolorbox}

\paragraph{Prompt for merge description attributes.}

We employ the following prompt to refine and polish the description attributes of an entity:

\begin{tcolorbox}[
  title=Merge Description Attributes,
  colback=white, 
  colframe=black,
  fonttitle=\bfseries,
  boxrule=0.5pt, 
  arc=2mm,
  left=2mm, 
  right=2mm, 
  top=1mm, 
  bottom=1mm,
  breakable
]
You are a medical AI assistant. Your task is to refine and polish the description of \{0\}.

The focus target is: ``\{0\}''.\\
The extracted description is: ``\{1\}''.\\
The chest X-ray radiological fact states: ``\{2\}''.

\textbf{Rules:}
\begin{enumerate}
    \item Correct the error or conflicting description to find the suitable description.
    \item Polish or ignore the unspecific description.
    \item Focus on the description of \{0\} only.
\end{enumerate}

\textbf{Output in the following format with <answer> and <think> tags:}\\
\texttt{<think>}\\
Thinking process to analyze and refine the content in one concise and short sentence; If no valid and correct description specified, clearly state this.\\
\texttt{</think>}\\
\texttt{<answer>}\\
Write the description in one sentence or paragraph from the view of diagnosis without mentioning the fact itself [Description]; If no valid and correct description specified, output ``None''.\\
\texttt{</answer>}

Stop immediately after outputting the results.
\end{tcolorbox}

\subsection{Report-level extraction}

In this phase, we extract \textbf{X‑ray quality assessment}, \textbf{findings relationships} and \textbf{diagnosis advice} from each radiology report; importantly, for every entity we also extract its pairwise relationships with all other entities, producing a more fine‑grained representation than a single global findings relationship. The prompt templates used to elicit these structured outputs from the LLM are as follows:

\paragraph{Prompt for relationships extraction.}

We employ the following prompt to extract the pathological relationships among findings:

\begin{tcolorbox}[
  title=Pathological Relationships Extraction,
  colback=white, 
  colframe=black,
  fonttitle=\bfseries,
  boxrule=0.5pt, 
  arc=2mm,
  left=2mm, 
  right=2mm, 
  top=1mm, 
  bottom=1mm,
  breakable
]
Please answer the following question as an X-ray assistant diagnosis with image, without referencing the report or mentioning yourself:\\
\textbf{Question}: ``What are the pathological relationships among the findings?''

The chest X-ray report: ``\{0\}''\\
Relationship words: ``\{1\}''\\
Findings: ``\{2\}''

\textbf{Rules:}
\begin{enumerate}
    \item Exclude entities without relationships or only with relationships to itself.
    \item Do not change the facts or infer relationships from context.
    \item Only the direct relationships between the findings are considered.
\end{enumerate}

\textbf{Output in the following format:}\\
\texttt{<think>} Analysis adhering to the rules and report in one sentence; If no relationships mentioned, explicitly state this. \texttt{</think>}

\texttt{<answer>}\\
Strictly using the report's language and detailed relationship words while summarizing the relationships among findings into one continuous and concise paragraph without listing and semicolon enumeration. If no relationships mentioned, please output ``None''.\\
\texttt{</answer>}

Stop immediately after outputting the results.
\end{tcolorbox}

\paragraph{Prompt for X-ray quality assessment extraction.}

We employ the following prompt to extract information regarding assessment difficulty, patient posture, image quality, and technical issues:

\begin{tcolorbox}[
  title=X-ray Quality Assessment Extraction,
  colback=white, 
  colframe=black,
  fonttitle=\bfseries,
  boxrule=0.5pt, 
  arc=2mm,
  left=2mm, 
  right=2mm, 
  top=1mm, 
  bottom=1mm,
  breakable
]
Please answer the following question as an X-ray assistant without referencing the report or mentioning itself:\\
\textbf{Question}: ``What are the X-ray issues regarding assessment difficulty, patient posture, image quality, and technical issues?''\\
Provide a one-sentence response based on your interpretation of the X-ray, or respond with ``None'' if no quality issues are mentioned.

The chest X-ray report: ``\{0\}''

\textbf{Rules:}
\begin{enumerate}
    \item Exclude any information about ``\{1\}''.
    \item Only include information related to assessment difficulty, patient posture, image quality, and technical issues.
    \item Do not infer issues from the report or change the facts if they are not explicitly mentioned.
\end{enumerate}

\textbf{Output in the following format:}\\
\texttt{<think>} Analysis adhering to the rules and report in one sentence; If no quality issues mentioned, explicitly state this. \texttt{</think>}

\texttt{<answer>}\\
A simple sentence that fully details the quality issue using the report's language, without duplication. If no quality issues mentioned, please output ``None''.\\
\texttt{</answer>}

Stop immediately after outputting the result.
\end{tcolorbox}

\paragraph{Prompt for diagnosis advice extraction.}

We employ the following prompt to extract  diagnosis advice for further imaging, clinical correlation, requirements, follow-up, uncertain diagnosis examination, or treatment:

\begin{tcolorbox}[
  title=Diagnosis Advice Extraction,
  colback=white, 
  colframe=black,
  fonttitle=\bfseries,
  boxrule=0.5pt, 
  arc=2mm,
  left=2mm, 
  right=2mm, 
  top=1mm, 
  bottom=1mm,
  breakable
]
Please answer the following question as an X-ray assistant without referencing the report or mentioning itself:\\
\textbf{Question}: ``What are the recommendations for further CT/MR/X-ray imaging, clinical correlation, requirements, follow-up, uncertain diagnosis examination, or treatment in this study?''\\
Provide a one-sentence response based on your interpretation of the X-ray, or respond with ``None'' if no recommendations are mentioned.

The chest X-ray report: ``\{0\}''

\textbf{Rules:}
\begin{enumerate}
    \item Focus on the recommendation information for further CT/MR/X-ray imaging, clinical correlation, requirements, follow-up, uncertain diagnosis examination, or treatment.
    \item If a CT/MR is mentioned in the report, it implies a CT/MR is recommended.
    \item Do not infer if it is not explicitly mentioned.
\end{enumerate}

\textbf{Output in the following format:}\\
\texttt{<think>} Analysis adhering to the rules and report in one sentence; If no recommendations mentioned, explicitly state this. \texttt{</think>}

\texttt{<answer>}\\
A simple sentence that fully details the recommendations using the report's language, without duplication. If no recommendations mentioned, please output ``None''.\\
\texttt{</answer>}

Stop immediately after outputting the result.
\end{tcolorbox}
\section{Limitations and Future Work}
\label{suppl:sec:limitation_and_future}

\paragraph{Extending modalities.}
While CXRTrekNet demonstrates strong performance on chest X-rays interpretation, its current architecture and training are primarily optimized for this modality. Consequently, direct application to other radiological modalities (e.g., CT, MRI) would likely require substantial adaptation. Future work will therefore focus on adapting the architecture and training methodologies for a broader range of imaging modalities. Furthermore, we plan to explore the integration of multimodal data sources, such as electronic health records (EHRs), to enrich contextual understanding and enhance the model's reasoning capabilities across diverse clinical situations.

\paragraph{Clinical validation and evaluation.}
The comprehensive evaluation of complex clinical reasoning presents an ongoing challenge. Although we employ clinical evaluation metrics like CheXbert-F1~\cite{smit2020chexbert} and Rad-Graph-F1~\cite{jain2021radgraph} for related tasks such as report generation, these may not fully capture the complete spectrum of diagnostic quality or the intricacies of human-like interpretation. Recognizing that robust clinical validation necessitates further collaboration with healthcare institutions, we plan to pursue such partnerships. We will also involve the incorporation of more nuanced evaluation metrics, aiming to more accurately reflect the sophisticated aspects of expert clinical interpretation.



\section{Broader impact}
\label{sec:broader_impact}



This research advances Medical Vision-Language Large Models (VLLMs) and enhances AI's understanding of medical context. We introduce a novel benchmark specifically designed for complex, multi-stage clinical reasoning, which emulates established clinical processes. This benchmark facilitates the development of more interpretable and reliable AI assistants. Such assistants promise to improve diagnostic efficiency and consistency, and to enrich medical education. Ultimately, this work aims to alleviate radiologist workload, enhance diagnostic precision, and elevate patient care by fostering more capable and clinically-aligned AI systems.


\newpage
\section*{NeurIPS Paper Checklist}
\begin{enumerate}

\item {\bf Claims}
    \item[] Question: Do the main claims made in the abstract and introduction accurately reflect the paper's contributions and scope?
    \item[] Answer: \answerYes{} 
    \item[] Justification: Yes, the abstract accurately reflects the paper's contributions: introducing the CXRTrek multi-stage VQA dataset to embed radiologists' sequential reasoning; proposing CXRTrekNet, a VLLM-based model capturing diagnostic stage connections and context; and demonstrating CXRTrekNet's superior benchmark performance and generalizability across five external datasets.

\item {\bf Limitations}
    \item[] Question: Does the paper discuss the limitations of the work performed by the authors?
    \item[] Answer: \answerYes{} 
    \item[] Justification: We have discussed the limitations of our work, and then further discussed the future direction of our work in the Appendix Section D.
    \item[] Guidelines:
    \begin{itemize}
        \item The answer NA means that the paper has no limitation while the answer No means that the paper has limitations, but those are not discussed in the paper. 
        \item The authors are encouraged to create a separate "Limitations" section in their paper.
        \item The paper should point out any strong assumptions and how robust the results are to violations of these assumptions ($e.g.$, independence assumptions, noiseless settings, model well-specification, asymptotic approximations only holding locally). The authors should reflect on how these assumptions might be violated in practice and what the implications would be.
        \item The authors should reflect on the scope of the claims made, $e.g.$, if the approach was only tested on a few datasets or with a few runs. In general, empirical results often depend on implicit assumptions, which should be articulated.
        \item The authors should reflect on the factors that influence the performance of the approach. For example, a facial recognition algorithm may perform poorly when image resolution is low or images are taken in low lighting. Or a speech-to-text system might not be used reliably to provide closed captions for online lectures because it fails to handle technical jargon.
        \item The authors should discuss the computational efficiency of the proposed algorithms and how they scale with dataset size.
        \item If applicable, the authors should discuss possible limitations of their approach to address problems of privacy and fairness.
        \item While the authors might fear that complete honesty about limitations might be used by reviewers as grounds for rejection, a worse outcome might be that reviewers discover limitations that aren't acknowledged in the paper. The authors should use their best judgment and recognize that individual actions in favor of transparency play an important role in developing norms that preserve the integrity of the community. Reviewers will be specifically instructed to not penalize honesty concerning limitations.
    \end{itemize}

\item {\bf Theory assumptions and proofs}
    \item[] Question: For each theoretical result, does the paper provide the full set of assumptions and a complete (and correct) proof?
    \item[] Answer: \answerNA{} 
    \item[] Justification: This paper does not propose new theories or theoretical results; hence, this item is not applicable. However, relevant existing theories or concepts that underpin the work are appropriately referenced and cited.

\item {\bf Experimental result reproducibility}
    \item[] Question: Does the paper fully disclose all the information needed to reproduce the main experimental results of the paper to the extent that it affects the main claims and/or conclusions of the paper (regardless of whether the code and data are provided or not)?
    \item[] Answer: \answerYes{}  
    \item[] Justification: The paper fully discloses all information needed to reproduce its main experimental results; these details are available in Section~\ref{sec:experiment} and are most comprehensively presented in Appendix Section A.
    \item[] Guidelines:
    \begin{itemize}
        \item The answer NA means that the paper does not include experiments.
        \item If the paper includes experiments, a No answer to this question will not be perceived well by the reviewers: Making the paper reproducible is important, regardless of whether the code and data are provided or not.
        \item If the contribution is a dataset and/or model, the authors should describe the steps taken to make their results reproducible or verifiable. 
        \item Depending on the contribution, reproducibility can be accomplished in various ways. For example, if the contribution is a novel architecture, describing the architecture fully might suffice, or if the contribution is a specific model and empirical evaluation, it may be necessary to either make it possible for others to replicate the model with the same dataset, or provide access to the model. In general. releasing code and data is often one good way to accomplish this, but reproducibility can also be provided via detailed instructions for how to replicate the results, access to a hosted model ($e.g.$, in the case of a large language model), releasing of a model checkpoint, or other means that are appropriate to the research performed.
        \item While NeurIPS does not require releasing code, the conference does require all submissions to provide some reasonable avenue for reproducibility, which may depend on the nature of the contribution. For example
        \begin{enumerate}
            \item If the contribution is primarily a new algorithm, the paper should make it clear how to reproduce that algorithm.
            \item If the contribution is primarily a new model architecture, the paper should describe the architecture clearly and fully.
            \item If the contribution is a new model ($e.g.$, a large language model), then there should either be a way to access this model for reproducing the results or a way to reproduce the model ($e.g.$, with an open-source dataset or instructions for how to construct the dataset).
            \item We recognize that reproducibility may be tricky in some cases, in which case authors are welcome to describe the particular way they provide for reproducibility. In the case of closed-source models, it may be that access to the model is limited in some way ($e.g.$, to registered users), but it should be possible for other researchers to have some path to reproducing or verifying the results.
        \end{enumerate}
    \end{itemize}

\item {\bf Open access to data and code}
    \item[] Question: Does the paper provide open access to the data and code, with sufficient instructions to faithfully reproduce the main experimental results, as described in supplemental material?
    \item[] Answer: \answerNo{} 
    \item[] Justification: Not open yet, but the paper commits to making all code, data (CXRTrek dataset), and model weights (CXRTrekNet model) openly accessible upon acceptance.
    \item[] Guidelines:
    \begin{itemize}
        \item The answer NA means that paper does not include experiments requiring code.
        \item Please see the NeurIPS code and data submission guidelines (\url{https://nips.cc/public/guides/CodeSubmissionPolicy}) for more details.
        \item While we encourage the release of code and data, we understand that this might not be possible, so “No” is an acceptable answer. Papers cannot be rejected simply for not including code, unless this is central to the contribution ($e.g.$, for a new open-source benchmark).
        \item The instructions should contain the exact command and environment needed to run to reproduce the results. See the NeurIPS code and data submission guidelines (\url{https://nips.cc/public/guides/CodeSubmissionPolicy}) for more details.
        \item The authors should provide instructions on data access and preparation, including how to access the raw data, preprocessed data, intermediate data, and generated data, etc.
        \item The authors should provide scripts to reproduce all experimental results for the new proposed method and baselines. If only a subset of experiments are reproducible, they should state which ones are omitted from the script and why.
        \item At submission time, to preserve anonymity, the authors should release anonymized versions (if applicable).
        \item Providing as much information as possible in supplemental material (appended to the paper) is recommended, but including URLs to data and code is permitted.
    \end{itemize}

\item {\bf Experimental setting/details}
    \item[] Question: Does the paper specify all the training and test details ($e.g.$, data splits, hyperparameters, how they were chosen, type of optimizer, etc.) necessary to understand the results?
    \item[] Answer: \answerYes{} 
    \item[] Justification: The paper provides detailed experimental settings in Section~\ref{sec:experiment} and the Appendix Section A.

\item {\bf Experiment statistical significance}
    \item[] Question: Does the paper report error bars suitably and correctly defined or other appropriate information about the statistical significance of the experiments?
    \item[] Answer:\answerNo{} 
    \item[] Justification: Error bars are not reported. However, we observe stable and consistent performance trends across all evaluation stages and datasets, which supports the reliability of our results.
    \item[] Guidelines:
    \begin{itemize}
        \item The answer NA means that the paper does not include experiments.
        \item The authors should answer "Yes" if the results are accompanied by error bars, confidence intervals, or statistical significance tests, at least for the experiments that support the main claims of the paper.
        \item The factors of variability that the error bars are capturing should be clearly stated (for example, train/test split, initialization, random drawing of some parameter, or overall run with given experimental conditions).
        \item The method for calculating the error bars should be explained (closed form formula, call to a library function, bootstrap, etc.)
        \item The assumptions made should be given ($e.g.$, Normally distributed errors).
        \item It should be clear whether the error bar is the standard deviation or the standard error of the mean.
        \item It is OK to report 1-sigma error bars, but one should state it. The authors should preferably report a 2-sigma error bar than state that they have a 96\% CI, if the hypothesis of Normality of errors is not verified.
        \item For asymmetric distributions, the authors should be careful not to show in tables or figures symmetric error bars that would yield results that are out of range ($e.g.$ negative error rates).
        \item If error bars are reported in tables or plots, The authors should explain in the text how they were calculated and reference the corresponding figures or tables in the text.
    \end{itemize}

\item {\bf Experiments compute resources}
    \item[] Question: For each experiment, does the paper provide sufficient information on the computer resources (type of compute workers, memory, time of execution) needed to reproduce the experiments?
    \item[] Answer: \answerYes{} 
    \item[] Justification: The paper provides detailed information about our computational resources in Section~\ref{sec:experiment}.
    \item[] Guidelines:
    \begin{itemize}
        \item The answer NA means that the paper does not include experiments.
        \item The paper should indicate the type of compute workers CPU or GPU, internal cluster, or cloud provider, including relevant memory and storage.
        \item The paper should provide the amount of compute required for each of the individual experimental runs as well as estimate the total compute. 
        \item The paper should disclose whether the full research project required more compute than the experiments reported in the paper ($e.g.$, preliminary or failed experiments that didn't make it into the paper). 
    \end{itemize}
    
\item {\bf Code of ethics}
    \item[] Question: Does the research conducted in the paper conform, in every respect, with the NeurIPS Code of Ethics \url{https://neurips.cc/public/EthicsGuidelines}?
    \item[] Answer: \answerYes{} 
    \item[] Justification: This research fully conforms to the NeurIPS Code of Ethics.
    \item[] Guidelines:
    \begin{itemize}
        \item The answer NA means that the authors have not reviewed the NeurIPS Code of Ethics.
        \item If the authors answer No, they should explain the special circumstances that require a deviation from the Code of Ethics.
        \item The authors should make sure to preserve anonymity ($e.g.$, if there is a special consideration due to laws or regulations in their jurisdiction).
    \end{itemize}

\item {\bf Broader impacts}
    \item[] Question: Does the paper discuss both potential positive societal impacts and negative societal impacts of the work performed?
    \item[] Answer: \answerYes{} 
    \item[] Justification: The broader impacts have been discussed in the Appendix Section E.
    \item[] Guidelines:
    \begin{itemize}
        \item The answer NA means that there is no societal impact of the work performed.
        \item If the authors answer NA or No, they should explain why their work has no societal impact or why the paper does not address societal impact.
        \item Examples of negative societal impacts include potential malicious or unintended uses ($e.g.$, disinformation, generating fake profiles, surveillance), fairness considerations ($e.g.$, deployment of technologies that could make decisions that unfairly impact specific groups), privacy considerations, and security considerations.
        \item The conference expects that many papers will be foundational research and not tied to particular applications, let alone deployments. However, if there is a direct path to any negative applications, the authors should point it out. For example, it is legitimate to point out that an improvement in the quality of generative models could be used to generate deepfakes for disinformation. On the other hand, it is not needed to point out that a generic algorithm for optimizing neural networks could enable people to train models that generate Deepfakes faster.
        \item The authors should consider possible harms that could arise when the technology is being used as intended and functioning correctly, harms that could arise when the technology is being used as intended but gives incorrect results, and harms following from (intentional or unintentional) misuse of the technology.
        \item If there are negative societal impacts, the authors could also discuss possible mitigation strategies ($e.g.$, gated release of models, providing defenses in addition to attacks, mechanisms for monitoring misuse, mechanisms to monitor how a system learns from feedback over time, improving the efficiency and accessibility of ML).
    \end{itemize}
    
\item {\bf Safeguards}
    \item[] Question: Does the paper describe safeguards that have been put in place for responsible release of data or models that have a high risk for misuse ($e.g.$, pretrained language models, image generators, or scraped datasets)?
    \item[] Answer: \answerYes{} 
    \item[] Justification: A core safeguard is that our CXRTrek dataset is constructed entirely from de-identified medical information, fundamentally ensuring data security and minimizing privacy risks. This responsible approach is reinforced by strict adherence to source data use agreements and a planned tiered access mechanism for dissemination.

\item {\bf Licenses for existing assets}
    \item[] Question: Are the creators or original owners of assets ($e.g.$, code, data, models), used in the paper, properly credited and are the license and terms of use explicitly mentioned and properly respected?
    \item[] Answer: \answerYes{} 
    \item[] Justification: Creators of all used assets are properly credited. Licenses and terms of use for these assets, such as foundational datasets (e.g., MIMIC-CXR) and pre-trained models (e.g., CheXagent), are explicitly mentioned and have been strictly respected.
    \item[] Guidelines:
    \begin{itemize}
        \item The answer NA means that the paper does not use existing assets.
        \item The authors should cite the original paper that produced the code package or dataset.
        \item The authors should state which version of the asset is used and, if possible, include a URL.
        \item The name of the license ($e.g.$, CC-BY 4.0) should be included for each asset.
        \item For scraped data from a particular source ($e.g.$, website), the copyright and terms of service of that source should be provided.
        \item If assets are released, the license, copyright information, and terms of use in the package should be provided. For popular datasets, \url{paperswithcode.com/datasets} has curated licenses for some datasets. Their licensing guide can help determine the license of a dataset.
        \item For existing datasets that are re-packaged, both the original license and the license of the derived asset (if it has changed) should be provided.
        \item If this information is not available online, the authors are encouraged to reach out to the asset's creators.
    \end{itemize}

\item {\bf New assets}
    \item[] Question: Are new assets introduced in the paper well documented and is the documentation provided alongside the assets?
    \item[] Answer: \answerNo{} 
    \item[] Justification: Our dataset (CXRTrek) and model (CXRTrekNet) are not yet released. Documentation will be provided with their release upon acceptance.

\item {\bf Crowdsourcing and research with human subjects}
    \item[] Question: For crowdsourcing experiments and research with human subjects, does the paper include the full text of instructions given to participants and screenshots, if applicable, as well as details about compensation (if any)? 
    \item[] Answer: \answerNA{} 
    \item[] Justification: The paper does not involve crowdsourcing nor research with human subjects.

\item {\bf Institutional review board (IRB) approvals or equivalent for research with human subjects}
    \item[] Question: Does the paper describe potential risks incurred by study participants, whether such risks were disclosed to the subjects, and whether Institutional Review Board (IRB) approvals (or an equivalent approval/review based on the requirements of your country or institution) were obtained?
    \item[] Answer: \answerNA{} 
    \item[] Justification: The paper does not involve crowdsourcing nor research with human subjects.

\item {\bf Declaration of LLM usage}
    \item[] Question: Does the paper describe the usage of LLMs if it is an important, original, or non-standard component of the core methods in this research? Note that if the LLM is used only for writing, editing, or formatting purposes and does not impact the core methodology, scientific rigorousness, or originality of the research, declaration is not required.
    \item[] Answer: \answerYes{} 
    \item[] Justification: The paper thoroughly describes the use of the relevant LLM in Sections~\ref{sec:dataset_construction}, \ref{sec:method}, and \ref{sec:experiment}. The LLM and its sources are correctly cited.

\end{enumerate}

\end{document}

%% file: assets/Tables/table_CXRTrek_compare_with_sota.tex
\begin{table*}[t!]
\centering
\small
\caption{Stage-wise performance when comparing our CXRTrekNet and advanced baselines on the CXRTrek test set. Metrics include BERTScore (BS) for open-ended responses, Macro-F1 score for closed-ended and choice responses, and mean IoU (mIoU) for detection (Det.) responses. `Ave.' means average performance over all stages. Metrics in \%.}
\setlength{\tabcolsep}{3.5pt}
\renewcommand{\arraystretch}{1.2}
\resizebox{1.0\columnwidth}{!}
{
\begin{tabular}{c|cc|ccc|cccc|c|c|c|c|c|c}
\toprule
\multirow{3}{*}{\textbf{Model}} & \multicolumn{2}{c|}{\textbf{Stage1}} & \multicolumn{3}{c|}{\textbf{Stage2}} & \multicolumn{4}{c|}{\textbf{Stage3}} & \textbf{Stage4} & \textbf{Stage5} & \textbf{Stage6} & \textbf{Stage7} & \textbf{Stage8} & \multirow{3}{*}{\textbf{Ave.}} \\ \cline{2-15}
& \multicolumn{1}{c|}{\textbf{Open}} & \textbf{Choice} & \multicolumn{1}{c|}{\textbf{Open}} & \multicolumn{1}{c|}{\textbf{Close}} & \textbf{Choice} & \multicolumn{1}{c|}{\textbf{Open}} & \multicolumn{1}{c|}{\textbf{Close}} & \multicolumn{1}{c|}{\textbf{Choice}} & \textbf{Det.} & \textbf{Open} & \textbf{Open} & \textbf{Open} & \textbf{Open} & \textbf{Open} & \\ \cline{2-15}
& \multicolumn{1}{c|}{BS}      & F1      & \multicolumn{1}{c|}{BS}      & \multicolumn{1}{c|}{F1}      & F1      & \multicolumn{1}{c|}{BS}      & \multicolumn{1}{c|}{F1}      & \multicolumn{1}{c|}{F1}      & mIoU      & BS      & BS      & BS      & BS      & BS      &                          \\ \hline
Qwen2-VL~\cite{wang2024qwen2}               & \multicolumn{1}{c|}{22.21} & 42.40 & \multicolumn{1}{c|}{22.74} & \multicolumn{1}{c|}{60.87} & 71.62 & \multicolumn{1}{c|}{23.27} & \multicolumn{1}{c|}{75.29} & \multicolumn{1}{c|}{63.01} & 0.36    & 30.53 & 25.59 & 15.99 & 17.98 & 20.94 & 35.20                  \\ \hline
XrayGPT~\cite{thawkar2023xraygpt}                & \multicolumn{1}{c|}{9.33}  & 24.23 & \multicolumn{1}{c|}{21.80} & \multicolumn{1}{c|}{50.60} & 29.98 & \multicolumn{1}{c|}{12.16} & \multicolumn{1}{c|}{44.70} & \multicolumn{1}{c|}{14.98} & 0.00    & 19.24 & 20.13 & 19.01 & 16.16 & 24.51 & 21.92                  \\ \hline
MiniGPT-Med~\cite{alkhaldi2024minigpt}            & \multicolumn{1}{c|}{12.27} & 36.69 & \multicolumn{1}{c|}{21.82} & \multicolumn{1}{c|}{40.58} & 40.50 & \multicolumn{1}{c|}{24.23} & \multicolumn{1}{c|}{59.40} & \multicolumn{1}{c|}{37.39} & 2.42    & 26.70 & 21.16 & 19.77 & 18.82 & 23.42 & 27.51                  \\ \hline
Llava-med v1.5~\cite{li2023llava}         & \multicolumn{1}{c|}{34.33} & 44.48 & \multicolumn{1}{c|}{23.03} & \multicolumn{1}{c|}{48.43} & 40.17 & \multicolumn{1}{c|}{22.85} & \multicolumn{1}{c|}{41.81} & \multicolumn{1}{c|}{49.94} & 1.24    & 28.20 & 16.69 & 13.67 & 16.46 & 10.26 & 27.97                  \\ \hline
CheXagent~\cite{chen2024chexagent}              & \multicolumn{1}{c|}{6.01}  & 33.25 & \multicolumn{1}{c|}{23.88} & \multicolumn{1}{c|}{76.66} & 46.69 & \multicolumn{1}{c|}{28.85} & \multicolumn{1}{c|}{65.16} & \multicolumn{1}{c|}{59.71} & 17.35   & 26.68 & 16.17 & 23.32 & 25.38 & 18.22 & 33.38                  \\ \hline
\rowcolor{lightgray} Our Model             & \multicolumn{1}{c|}{\textbf{56.02}} & \textbf{98.21} & \multicolumn{1}{c|}{\textbf{62.48}} & \multicolumn{1}{c|}{\textbf{77.36}} & \textbf{71.84} & \multicolumn{1}{c|}{\textbf{61.13}} & \multicolumn{1}{c|}{\textbf{90.11}} & \multicolumn{1}{c|}{\textbf{96.26}} & \textbf{49.65}   & \textbf{66.60} & \textbf{72.33} & \textbf{69.48} & \textbf{53.21} & \textbf{50.52} & \textbf{69.66}                  \\ \bottomrule
\end{tabular}
}
\label{tab:stagewise_results}
\end{table*}

%% file: assets/Tables/table_vqa_and_classification.tex
\begin{table}[t!]
  \centering
  \setlength{\tabcolsep}{5pt}
  \footnotesize 
  \caption{Comparison of model performance across the VQA‑RAD, Rad-ReStruct, VinDr‑CXR datasets. Following the evaluation protocol of LLaVA-Med for VQA-RAD, open-ended questions are evaluated using recall, while closed-ended questions are assessed using accuracy. For Rad-ReStruct, we report individual Q\&A accuracy (Acc) and full report accuracy (Report Acc). For VinDR-CXR classification tasks, both accuracy (Acc) and Macro-F1 (F1) score are used as evaluation metrics.} 
  \label{tab:perf_combined} 

  \begin{minipage}[t]{0.58\textwidth} 
    \vspace{0pt}
    \centering 
    \begin{tabular}{lcccc} 
    \toprule
    \multirow{2}{*}{\textbf{Model}}
      & \multicolumn{2}{c}{\textbf{VQA‑RAD}}
      & \multicolumn{2}{c}{\textbf{Rad-ReStruct}} \\
    \cmidrule(lr){2-3}\cmidrule(lr){4-5} 
      & Open & Close & Report Acc & Acc  \\
    \midrule
    LLaVA‑Med v1.5 \cite{li2023llava}  & 32.17 & 65.07  &  0.00  &  37.59 \\
    MiniGPT‑Med   \cite{alkhaldi2024minigpt}  & 48.73 & 77.57  &  36.37 & 85.67   \\
    CheXagent     \cite{chen2024chexagent}   & 19.95 & 68.75  &  22.22 &  77.59   \\
    \rowcolor{lightgray} 
    Our Model     & \textbf{54.50} & \textbf{82.35} &  \textbf{38.60}  &  \textbf{91.28}  \\
    \bottomrule
    \end{tabular}
  \end{minipage}
  \hfill 
  \begin{minipage}[t]{0.38\textwidth} 
    \vspace{0pt}
    \centering 
    \begin{tabular}{lcc} 
    \toprule
    \multirow{2}{*}{\textbf{Model}}
      & \multicolumn{2}{c}{\textbf{VinDr‑CXR}} \\
    \cmidrule(lr){2-3} 
    & Acc & F1  \\
    \midrule
    LLaVA‑Med v1.5 \cite{li2023llava}   & 46.9  & 6.91 \\
    MiniGPT‑Med   \cite{alkhaldi2024minigpt}  & 92.87 & 3.72   \\
    CheXagent     \cite{chen2024chexagent}    & 90.41 & 30.18   \\
    \rowcolor{lightgray} 
    Our Model     & \textbf{96.06} & \textbf{33.35} \\
    \bottomrule
    \end{tabular}
  \end{minipage}
\end{table}

%% file: assets/Tables/table_mimic_cxr.tex
\begin{table*}[t!]
\centering
\footnotesize
\caption{Comparison of experimental results for models on the report generation task. Natural language generation metrics include R-L (ROUGE-L) and BS (BERTScore~\cite{zhang2019bertscore}). Clinical evaluation metrics include C-F1 (CheXbert-F1~\cite{smit2020chexbert}) and R-F1 (RadGraph-F1~\cite{jain2021radgraph}). Metrics in \%.}
\label{tab:report_generation}
\setlength{\tabcolsep}{5pt} 

\begin{tabularx}{\linewidth}{l ccccc | ccccc}
\toprule
\multirow{2}{*}{\textbf{Model}} & \multicolumn{5}{c}{\textbf{MIMIC-CXR}} & \multicolumn{5}{c}{\textbf{IU-Xray}} \\
\cmidrule(lr){2-6} \cmidrule(lr){7-11}
& R-L & BS & C-F1 & R-F1 & Mean & R-L & BS & C-F1 & R-F1 & Mean \\
\midrule
LLaVA-Med v1.5 \cite{li2023llava} & 16.92 & 26.1 & 16.66  & 4.88 &  16.14 & 16.01 & 21.12 & 13.96 & 7.49 & 14.64 \\
MiniGPT-Med  \cite{alkhaldi2024minigpt}  & 27.03  & 36.08  & 21.42 & 14.72 & 24.81 & 34.55 & 47.26& 48.84 & 30.68 & 40.33 \\
XrayGPT \cite{thawkar2023xraygpt} & 28.43 & 36.74 & 29.02 & 16.35 & 27.63  & 30.89  & 42.04  & 28.24 & 22.69  & 30.96  \\
CheXagent \cite{chen2024chexagent} & 33.59 & 31.74 & \textbf{47.63} & \textbf{23.52} & 34.12 & 32.17 & 39.34 &  46.17  &  24.81  & 35.62 \\
\rowcolor{lightgray}
Our Model   & \textbf{37.49} & \textbf{40.34} & 43.27  & 21.96 & \textbf{35.76}  & \textbf{40.75}  & \textbf{52.53}  & \textbf{49.00} & \textbf{31.44} & \textbf{43.43} \\
\bottomrule
\end{tabularx}
\end{table*}

%% file: assets/Tables/table_CXRTrek_Ablation.tex
\begin{table*}[t!]
\centering
\small
\caption{Ablation studies on the CXRTrek test set. Metrics include BERTScore (BS) for open-ended responses, Macro-F1 score for closed-ended and choice responses, and mean IoU (mIoU) for detection (Det.) responses. `Joint' refers to combining Q\&A pairs from all stages into a single-stage training process. `Ours' denotes sequential stage-wise training guided by contextually clinical reasoning. `x\%' indicates the proportion of training data used. For a fair comparison, all models were trained for 5k iterations using identical parameter configurations. Metrics in \%.}
\setlength{\tabcolsep}{3.5pt}
\renewcommand{\arraystretch}{1.2}
\resizebox{1.0\columnwidth}{!}
{
\begin{tabular}{c|cc|ccc|cccc|c|c|c|c|c|c}
\toprule
\multirow{3}{*}{\textbf{Model}} & \multicolumn{2}{c|}{\textbf{Stage1}} & \multicolumn{3}{c|}{\textbf{Stage2}} & \multicolumn{4}{c|}{\textbf{Stage3}} & \textbf{Stage4} & \textbf{Stage5} & \textbf{Stage6} & \textbf{Stage7} & \textbf{Stage8} & \multirow{3}{*}{\textbf{Ave.}} \\ \cline{2-15}
& \multicolumn{1}{c|}{\textbf{Open}} & \textbf{Choice} & \multicolumn{1}{c|}{\textbf{Open}} & \multicolumn{1}{c|}{\textbf{Close}} & \textbf{Choice} & \multicolumn{1}{c|}{\textbf{Open}} & \multicolumn{1}{c|}{\textbf{Close}} & \multicolumn{1}{c|}{\textbf{Choice}} & \textbf{Det.} & \textbf{Open} & \textbf{Open} & \textbf{Open} & \textbf{Open} & \textbf{Open} & \\ \cline{2-15}
                        & \multicolumn{1}{c|}{BS}      & F1      & \multicolumn{1}{c|}{BS}      & \multicolumn{1}{c|}{F1}      & F1      & \multicolumn{1}{c|}{BS}      & \multicolumn{1}{c|}{F1}      & \multicolumn{1}{c|}{F1}      & mIoU    & BS      & BS      & BS      & BS      & BS      &                       \\ \hline
Joint (100\%)          & \multicolumn{1}{c|}{54.11} & 41.85 & \multicolumn{1}{c|}{45.46} & \multicolumn{1}{c|}{64.83} & 62.89 & \multicolumn{1}{c|}{50.58} & \multicolumn{1}{c|}{73.32} & \multicolumn{1}{c|}{76.96} & 25.37 & 50.13 & 57.57 & 62.51 & 45.67 & 28.86 & 52.86               \\ \hline
Ours (1\%)           & \multicolumn{1}{c|}{\textbf{55.02}} & 49.74 & \multicolumn{1}{c|}{55.69} & \multicolumn{1}{c|}{\textbf{72.72}} & 64.57 & \multicolumn{1}{c|}{54.24} & \multicolumn{1}{c|}{79.05} & \multicolumn{1}{c|}{\textbf{93.70}} & 36.33 & 58.69 & 64.56 & 66.77 & 50.21 & 38.80 & 60.01               \\ \hline
Ours (100\%)         & \multicolumn{1}{c|}{54.19} & \textbf{81.92} & \multicolumn{1}{c|}{\textbf{60.07}} & \multicolumn{1}{c|}{71.53} & \textbf{65.85} & \multicolumn{1}{c|}{\textbf{58.67}} & \multicolumn{1}{c|}{\textbf{82.14}} & \multicolumn{1}{c|}{92.20} & \textbf{41.47} & \textbf{62.76} & \textbf{67.79} & \textbf{70.67} & \textbf{50.68} & \textbf{45.86} & \textbf{64.70}               \\ \bottomrule
\end{tabular}
}
\label{tab:ablation}
\end{table*}

%% file: assets/Tables/table_st_mt_on_cxrtrek.tex
\begin{table*}[h!]
\centering
\small
\caption{Stage-wise performance comparison of our model and advanced baselines on the CXRTrek test set. Metrics include BERTScore~\cite{zhang2019bertscore} (BS) for open-ended responses, Macro-F1 score for closed-ended and multiple-choice responses, and mean Intersection over Union (mIoU) for detection (Det.) tasks. 'Ave.' denotes average performance across all stages. All scores are presented as percentages (\%). The highlighted rows indicate the primary results selected for comparison for each model.}
\setlength{\tabcolsep}{3.5pt}
\renewcommand{\arraystretch}{1.2}
\resizebox{1.0\columnwidth}{!}{
\begin{tabular}{c|c|cc|ccc|cccc|c|c|c|c|c|c}
\toprule
\multirow{3}{*}{\textbf{Model}} & \multirow{3}{*}{\makecell[c]{\textbf{Test} \\ \textbf{Strat.}}} & \multicolumn{2}{c|}{\textbf{Stage1}} & \multicolumn{3}{c|}{\textbf{Stage2}} & \multicolumn{4}{c|}{\textbf{Stage3}} & \multicolumn{1}{c|}{\textbf{Stage4}} & \multicolumn{1}{c|}{\textbf{Stage5}} & \multicolumn{1}{c|}{\textbf{Stage6}} & \multicolumn{1}{c|}{\textbf{Stage7}} & \multicolumn{1}{c|}{\textbf{Stage8}} & \multirow{3}{*}{\textbf{Ave.}} \\ 
\cline{3-16}
& & \multicolumn{1}{c|}{\textbf{Open}} & \textbf{Choice} & \multicolumn{1}{c|}{\textbf{Open}} & \multicolumn{1}{c|}{\textbf{Close}} & \textbf{Choice} & \multicolumn{1}{c|}{\textbf{Open}} & \multicolumn{1}{c|}{\textbf{Close}} & \multicolumn{1}{c|}{\textbf{Choice}} & \textbf{Det.} & \textbf{Open} & \textbf{Open} & \textbf{Open} & \textbf{Open} & \textbf{Open} & \\ 
\cline{3-16}
& & \multicolumn{1}{c|}{BS} & F1 & \multicolumn{1}{c|}{BS} & \multicolumn{1}{c|}{F1} & F1 & \multicolumn{1}{c|}{BS} & \multicolumn{1}{c|}{F1} & \multicolumn{1}{c|}{F1} & mIoU & BS & BS & BS & BS & BS & \\ 
\midrule
\multirow{2}{*}{Qwen2-VL~\cite{wang2024qwen2}} 
  & \cellcolor{lightgray}Joint & \multicolumn{1}{c|}{\cellcolor{lightgray}22.21} & \cellcolor{lightgray}42.40 & \multicolumn{1}{c|}{\cellcolor{lightgray}22.74} & \multicolumn{1}{c|}{\cellcolor{lightgray}60.87} & \cellcolor{lightgray}71.62 & \multicolumn{1}{c|}{\cellcolor{lightgray}23.27} & \multicolumn{1}{c|}{\cellcolor{lightgray}75.29} & \multicolumn{1}{c|}{\cellcolor{lightgray}63.01} & \cellcolor{lightgray}0.36 & \cellcolor{lightgray}30.53 & \cellcolor{lightgray}25.59 & \cellcolor{lightgray}15.99 & \cellcolor{lightgray}17.98 & \cellcolor{lightgray}20.94 & \cellcolor{lightgray}35.20 \\
  & Multi-Stage  & \multicolumn{1}{c|}{21.77} & 33.58 & \multicolumn{1}{c|}{23.85} & \multicolumn{1}{c|}{55.73} & 50.62 & \multicolumn{1}{c|}{23.49} & \multicolumn{1}{c|}{67.15} & \multicolumn{1}{c|}{73.39} & 0.30 & 28.77 & 22.73 & 18.95 & 18.75 & 21.75 & 32.92 \\ 
\midrule
\multirow{2}{*}{XrayGPT~\cite{thawkar2023xraygpt}}   & 
 \cellcolor{lightgray}Joint  & \multicolumn{1}{c|}{\cellcolor{lightgray}9.33}  & \cellcolor{lightgray}24.23 & \multicolumn{1}{c|}{\cellcolor{lightgray}21.80} & \multicolumn{1}{c|}{\cellcolor{lightgray}50.60} & \cellcolor{lightgray}29.98 & \multicolumn{1}{c|}{\cellcolor{lightgray}12.16} & \multicolumn{1}{c|}{\cellcolor{lightgray}44.70} & \multicolumn{1}{c|}{\cellcolor{lightgray}14.98} & \cellcolor{lightgray}0.00 & \cellcolor{lightgray}19.24 & \cellcolor{lightgray}20.13 & \cellcolor{lightgray}19.01 & \cellcolor{lightgray}16.16 & \cellcolor{lightgray}24.51 & \cellcolor{lightgray}21.92 \\
  & Multi-Stage  & \multicolumn{1}{c|}{8.96}  & 13.13 & \multicolumn{1}{c|}{15.20} & \multicolumn{1}{c|}{45.76} & 26.14 & \multicolumn{1}{c|}{4.62}  & \multicolumn{1}{c|}{32.98} & \multicolumn{1}{c|}{30.56} & 0.00 & -17.70 & -7.10 & -11.48 & -15.18 & -15.48 & 7.89 \\ 
\midrule
\multirow{2}{*}{MiniGPT-Med~\cite{alkhaldi2024minigpt}}
& \cellcolor{lightgray}Joint  & \multicolumn{1}{c|}{\cellcolor{lightgray}12.27} & \cellcolor{lightgray}36.69 & \multicolumn{1}{c|}{\cellcolor{lightgray}21.82} & \multicolumn{1}{c|}{\cellcolor{lightgray}40.58} & \cellcolor{lightgray}40.50 & \multicolumn{1}{c|}{\cellcolor{lightgray}24.23} & \multicolumn{1}{c|}{\cellcolor{lightgray}59.40} & \multicolumn{1}{c|}{\cellcolor{lightgray}37.39} & \cellcolor{lightgray}2.42 & \cellcolor{lightgray}26.70 & \cellcolor{lightgray}21.16 & \cellcolor{lightgray}19.77 & \cellcolor{lightgray}18.82 & \cellcolor{lightgray}23.42 & \cellcolor{lightgray}27.51 \\
  & Multi-Stage  & \multicolumn{1}{c|}{9.89}  & 31.65 & \multicolumn{1}{c|}{19.22} & \multicolumn{1}{c|}{44.58} & 29.51 & \multicolumn{1}{c|}{20.46} & \multicolumn{1}{c|}{55.67} & \multicolumn{1}{c|}{24.42} & 1.06 & 7.96  & 18.30 & 14.68 & 15.94 & 17.22 & 22.18 \\ 
\midrule
\multirow{2}{*}{Llava-med v1.5~\cite{li2023llava}}  & \cellcolor{lightgray}Joint  & \multicolumn{1}{c|}{\cellcolor{lightgray}34.33} & \cellcolor{lightgray}44.48 & \multicolumn{1}{c|}{\cellcolor{lightgray}23.03} & \multicolumn{1}{c|}{\cellcolor{lightgray}48.43} & \cellcolor{lightgray}40.17 & \multicolumn{1}{c|}{\cellcolor{lightgray}22.85} & \multicolumn{1}{c|}{\cellcolor{lightgray}41.81} & \multicolumn{1}{c|}{\cellcolor{lightgray}49.94} & \cellcolor{lightgray}1.24 & \cellcolor{lightgray}28.20 & \cellcolor{lightgray}16.69 & \cellcolor{lightgray}13.67 & \cellcolor{lightgray}16.46 & \cellcolor{lightgray}10.26 & \cellcolor{lightgray}27.97 \\
  & Multi-Stage  & \multicolumn{1}{c|}{33.88} & 38.74 & \multicolumn{1}{c|}{16.75} & \multicolumn{1}{c|}{51.07} & 51.35 & \multicolumn{1}{c|}{12.02} & \multicolumn{1}{c|}{30.98} & \multicolumn{1}{c|}{41.58} & 1.19 & -0.86 & -4.10 & -1.76 & -3.10 & -8.27 & 18.53 \\ 
\midrule
\multirow{2}{*}{CheXagent~\cite{chen2024chexagent}} 
 &\cellcolor{lightgray}  Joint  & \multicolumn{1}{c|}{\cellcolor{lightgray}6.01}  & \cellcolor{lightgray}33.25 & \multicolumn{1}{c|}{\cellcolor{lightgray}23.88} & \multicolumn{1}{c|}{\cellcolor{lightgray}76.66} & \cellcolor{lightgray}46.69 & \multicolumn{1}{c|}{\cellcolor{lightgray}28.85} & \multicolumn{1}{c|}{\cellcolor{lightgray}65.16} & \multicolumn{1}{c|}{\cellcolor{lightgray}59.71} & \cellcolor{lightgray}17.35 & \cellcolor{lightgray}26.68 & \cellcolor{lightgray}16.17 & \cellcolor{lightgray}23.32 & \cellcolor{lightgray}25.38 & \cellcolor{lightgray}18.22 & \cellcolor{lightgray}33.38 \\
  & Multi-Stage  & \multicolumn{1}{c|}{1.93}  & 36.13 & \multicolumn{1}{c|}{-4.84} & \multicolumn{1}{c|}{62.18} & 47.14 & \multicolumn{1}{c|}{11.59} & \multicolumn{1}{c|}{32.65} & \multicolumn{1}{c|}{48.64} & 5.45  & 9.77  & -10.37& 7.91  & 4.88  & 7.00  & 18.58 \\ 
\midrule
\multirow{2}{*}{Our Model} 
  & Joint        & \multicolumn{1}{c|}{54.05} & 98.21 & \multicolumn{1}{c|}{53.19} & \multicolumn{1}{c|}{82.88} & 70.90 & \multicolumn{1}{c|}{55.86} & \multicolumn{1}{c|}{81.82} & \multicolumn{1}{c|}{93.09} & 47.63 & 61.29 & 62.86 & 65.07 & 46.97 & 33.82 & 64.83 \\ 
 & \cellcolor{lightgray}Multi-Stage  & \multicolumn{1}{c|}{\cellcolor{lightgray}\textbf{56.02}} & \cellcolor{lightgray}\textbf{98.21} & \multicolumn{1}{c|}{\cellcolor{lightgray}\textbf{62.48}} & \multicolumn{1}{c|}{\cellcolor{lightgray}\textbf{77.36}} & \cellcolor{lightgray}\textbf{71.84} & \multicolumn{1}{c|}{\cellcolor{lightgray}\textbf{61.13}} & \multicolumn{1}{c|}{\cellcolor{lightgray}\textbf{90.11}} & \multicolumn{1}{c|}{\cellcolor{lightgray}\textbf{96.26}} & \cellcolor{lightgray}\textbf{49.65} & \cellcolor{lightgray}\textbf{66.60} & \cellcolor{lightgray}\textbf{72.33} & \cellcolor{lightgray}\textbf{69.48} & \cellcolor{lightgray}\textbf{53.21} & \cellcolor{lightgray}\textbf{50.52} & \cellcolor{lightgray}\textbf{69.66} \\
\bottomrule
\end{tabular}}
\label{tab:st_mt_cxrtrek_results}
\end{table*}

%% file: assets/Tables/table_evidence_from_radpapers.tex
\begin{table}[ht]
  \centering
  \scriptsize
  \caption{Eight stages of dataset processing with corresponding standard-guide evidence}
  \label{tab:table_evidence_from_radpapers}
  \begin{tabularx}{\textwidth}{@{}
    >{\RaggedRight\arraybackslash}p{0.25\textwidth}
    >{\RaggedRight\arraybackslash}p{0.7\textwidth}
  @{}}
    \toprule
    \textbf{8 Stages of Our Dataset} & \textbf{Supporting Evidence from Standard Guide} \\
    \midrule
    \parbox[t]{\linewidth}{Stage 1: X-ray Validation} &
    \parbox[t]{\linewidth}{The initial evaluation of any chest radiograph should include a determination of the technical adequacy of the examination to confirm that it is of adequate quality for interpretation. This step is often overlooked, which can lead to both overdiagnosis and underdiagnosis. ~\cite{klein2019systematic}} \\
    \midrule
    \parbox[t]{\linewidth}{Stage 2: Findings Recognition} &
    \parbox[t]{\linewidth}{The orderly assessment of each anatomic region and structure will yield a comprehensive imaging evaluation, will allow identification of subtle abnormalities, and will minimize interpretive errors. The following must be evaluated in each chest radiograph: support and monitoring devices; chest wall; heart and mediastinum; hila; lungs; airways; pleura; diaphragm. \cite{klein2019systematic}} \\
    \midrule
    \parbox[t]{\linewidth}{Stage 3: Findings’ Attributes Recognition} &
    \parbox[t]{\linewidth}{Key features that should be assessed for pulmonary nodules include morphology, density, growth, and metabolic activity. \cite{klein2019systematic}} \\
    \midrule
    \parbox[t]{\linewidth}{Stage 4: Findings’ Relationships Questions} &
    \parbox[t]{\linewidth}{The differential diagnosis may be narrowed by referring to a combination of imaging features, disease chronicity or progression, response to treatment, and the immune status of the patient. \cite{klein2019systematic}} \\
    \midrule
    \parbox[t]{\linewidth}{Stage 5: Image Comparisons} &
    \parbox[t]{\linewidth}{Every effort should be made to review previous chest radiographs and their reports, thereby providing one of the best learning tools for chest radiograph interpretation. \cite{klein2019systematic}} \\
    \midrule
    \parbox[t]{\linewidth}{Stage 6: Future Risk Questions} &
    \parbox[t]{\linewidth}{The identification of diffuse, central, lamellated, or popcorn-type calcification in a pulmonary nodule is diagnostic of a benign lesion. Eccentric calcification is indeterminate—sometimes benign and related to granulomatous disease—but also occurring in malignant lesions such as lung cancer or a carcinoid tumor. The presence of calcification can also be difficult to discern on standard high-kVp chest radiographs. \cite{gelaw2015screening}} \\
    \midrule
    \parbox[t]{\linewidth}{Stage 7: Diagnosis Advice} &
    \parbox[t]{\linewidth}{Standard CXR interpretation requires a systematic approach including the following steps: checking the identification of the image; evaluating the image quality; systematically reviewing the CXR for abnormalities; detecting and describing the abnormalities; suggesting a generic process, and whenever possible, a specific process that might explain the abnormality; and recommending other imaging modalities if needed. \cite{gelaw2015screening}\\
    If the radiologist cannot arrive at a conclusive diagnosis from the projections that the technologist provides, he or she must recommend other imaging procedures or follow-up projections. \cite{boecking2022ms}} \\
    \midrule
    \parbox[t]{\linewidth}{Stage 8: Summarizing} &
    \parbox[t]{\linewidth}{All diagnostic information must be recorded and entered into the patient's electronic medical record system. \cite{boecking2022ms}} \\
    \bottomrule
  \end{tabularx}
\end{table}

%% file: assets/Tables/table_corresponding_tasks.tex
\begin{table}[t]
  \centering
  \scriptsize
  \caption{Comparison of Our Q\&A Task Types}
  \label{tab:table_corresponding_tasks}
  \begin{tabularx}{\textwidth}{@{}
    >{\RaggedRight\arraybackslash}p{0.06\textwidth}
    >{\RaggedRight\arraybackslash}p{0.12\textwidth}
    >{\RaggedRight\arraybackslash}p{0.22\textwidth}
    >{\RaggedRight\arraybackslash}p{0.14\textwidth}
    >{\RaggedRight\arraybackslash}p{0.12\textwidth}
    >{\RaggedRight\arraybackslash}p{0.20\textwidth}
  @{}}
    \toprule
    \textbf{Stage} & 
    \textbf{\makecell[l]{Task Type of \\ question}} & \textbf{Examples in Our Dataset} & \textbf{Other Datasets} & \textbf{Corresponding Task Types} & \textbf{Examples in Other Datasets} \\
    \midrule
    \multirow{2}{*}{Stage-1} & View Information &
    \parbox[t]{\linewidth}{Describe the orientation of this X-ray view.} &
    \parbox[t]{\linewidth}{VQA-Med~\cite{ben2019vqa};\\VQA-RAD~\cite{lau2018dataset};\\SLAKE~\cite{liu2021slake};\\Medical-CXR-VQA~\cite{humedical};\\Rad-Restruct~\cite{pellegrini2023rad};\\CheXbench~\cite{chen2024chexagent}} &
    \parbox[t]{\linewidth}{Plane;\\View;\\Gender} &
    \parbox[t]{\linewidth}{In what plane was this image taken?} \\
    \cmidrule{2-6}
    & \parbox[t]{\linewidth}{Quality \\Assessment} &
    \parbox[t]{\linewidth}{How is the quality of the image?} &
    \parbox[t]{\linewidth}{-} &
    \parbox[t]{\linewidth}{-} &
    \parbox[t]{\linewidth}{-} \\
    \midrule
    \multirow{3}{*}{Stage-2} & \parbox[t]{\linewidth}{Findings \\ Identification} &
    \parbox[t]{\linewidth}{What abnormalities or foreign objects can be found in the X-ray(s)?\\Are there any abnormalities in this image?} &
    \parbox[t]{\linewidth}{VQA-Med~\cite{ben2019vqa};\\VQA-RAD~\cite{lau2018dataset};\\SLAKE~\cite{liu2021slake};\\Medical-CXR-VQA~\cite{humedical};\\VinDr-CXR~\cite{nguyen2022vindr}; \\Rad-Restruct~\cite{pellegrini2023rad};\\CheXbench~\cite{chen2024chexagent} } &
    \parbox[t]{\linewidth}{abnormality;\\ABN;\\PRES;\\Organ;\\Presence} &
    \parbox[t]{\linewidth}{Are there abnormalities in this image?} \\
    \midrule
    \multirow{2}{*}{Stage-3} & \parbox[t]{\linewidth}{Attributes \\Description} &
    \parbox[t]{\linewidth}{How severe is \{object\}?\\Where is the location of \{object\}?\\What type is the \{object\}?\\How would you describe the \{object\}?} &
    \parbox[t]{\linewidth}{VQA-RAD~\cite{lau2018dataset};\\SLAKE~\cite{liu2021slake};\\Medical-CXR-VQA~\cite{humedical};\\Rad-Restruct~\cite{pellegrini2023rad}} &
    \parbox[t]{\linewidth}{POS;\\Size;\\ATTRIB;\\Position;\\Quantity;\\Location;\\Type;\\Level} &
    \parbox[t]{\linewidth}{Where is there an obstruction present?} \\
    \cmidrule{2-6}
   & Phrase Grounding; / Findings Detection&
    \parbox[t]{\linewidth}{Locate bounding box(es) for the phrase: \{phrase\}. \\ Draw the bounding box of \{object\}.} &
    \parbox[t]{\linewidth}{MS-CXR~\cite{boecking2022ms};\\NLST~\cite{national2011national};\\CheXbench~\cite{chen2024chexagent}; \\ RSNA~\cite{rsna-pneumonia-detection-challenge}; \\ VinDr-CXR~\cite{nguyen2022vindr}; \\ Chest-imaGenome~\cite{wu2021chest}} &
    \parbox[t]{\linewidth}{Grounding; \\ Detection} &
    \parbox[t]{\linewidth}{Locate the area for the phrase: \{phrase\}.} \\
    \midrule
  \multirow{1}{*}{Stage-4} & Relationships Analysis &
    \parbox[t]{\linewidth}{Describe the relationships among \{object\} in the X-ray image.\\How does \{object\} influence \{object2\} in the X-ray image?} &
    \parbox[t]{\linewidth}{-} &
    \parbox[t]{\linewidth}{-} &
    \parbox[t]{\linewidth}{-} \\
    \midrule
   \multirow{1}{*}{Stage-5} & \parbox[t]{\linewidth}{ Differences \\Comparison} &
    \parbox[t]{\linewidth}{What differences can be observed in the findings between the current and previous images?} &
    \parbox[t]{\linewidth}{MS-CXR-T~\cite{bannur2023ms};\\MIMIC-Diff-VQA~\cite{hu2023expert};\\CheXbench~\cite{chen2024chexagent}} &
       \parbox[t]{\linewidth}{Differences Comparison;\\Temporal Classification} &
    \parbox[t]{\linewidth}{Given two CXRs, decide if pleural effusion (A) has improved, (B) is stable, or (C) has worsened.} \\
    \midrule
    \multirow{1}{*}{Stage-6} & Future Trend Analysis &
    \parbox[t]{\linewidth}{What are the trends and potential risks of the abnormalities in the current X-ray image?} &
    \parbox[t]{\linewidth}{-} &
    \parbox[t]{\linewidth}{-} &
    \parbox[t]{\linewidth}{-} \\
    \midrule
    \multirow{1}{*}{Stage-7} &  Diagnosis Advice &
    \parbox[t]{\linewidth}{What diagnostic advice would you provide to the referring physician based on the X-ray findings?} &
    \parbox[t]{\linewidth}{-} &
    \parbox[t]{\linewidth}{-} &
    \parbox[t]{\linewidth}{-} \\
    \midrule
    \multirow{1}{*}{Stage-8}  & \parbox[t]{\linewidth}{Report/Findinds/\\Impression \\Generation} &
     \parbox[t]{\linewidth}{Write a report/findings section or impression section based on the X-ray image.} &
     \parbox[t]{\linewidth}{IU-Xray~\cite{demner2016preparing};\\MIMIC-CXR~\cite{johnson2019mimic};\\Chexpert-plus~\cite{chambon2024chexpert};\\CheXbench~\cite{chen2024chexagent}} &
     \parbox[t]{\linewidth}{Findings Generation;\\Report Generation;\\Findings Summarization} &
     \parbox[t]{\linewidth}{Write its Findings section;\\Write its Impression section.} \\
    \bottomrule
  \end{tabularx}
\end{table}

%% file: assets/Tables/table_check_extract.tex
\begin{table}[h!]
  \centering
  \footnotesize 
  \setlength{\tabcolsep}{2.5pt} 
  \caption{
  Quantitative evaluation of model-extracted labels against human annotations using BERTScore across three diagnostic dimensions: (1) Entities (Abn.: abnormalities, Fgn.: foreign objects), (2) Attributes (Loc.: location, Sev.: severity, Trait: characteristics, Trend: progression, Desc.: description), and (3) Diagnostic Context (Qual.: X-ray quality assessment, Rel.: findings’ relationships, Adv.: diagnosis advice). The overall high scores indicate strong semantic alignment and annotation quality.}
  \label{tab:diagnosis_context}
  \begin{tabular}{
    >{\raggedright\arraybackslash}p{0.1\linewidth} 
    *{2}{>{\centering\arraybackslash}p{0.068\linewidth}} 
    *{5}{>{\centering\arraybackslash}p{0.068\linewidth}} 
    *{3}{>{\centering\arraybackslash}p{0.068\linewidth}} 
    >{\centering\arraybackslash}p{0.06\linewidth} 
  }
    \toprule
    & \multicolumn{2}{c}{\textbf{Entities}}
    & \multicolumn{5}{c}{\textbf{Attributes}}
    & \multicolumn{3}{c}{\textbf{Diagnosis Context}} 
    & \multirow{2}{*}{\textbf{Mean}}\\
    \cmidrule(lr){2-3} \cmidrule(lr){4-8} \cmidrule(lr){9-11}
    \textbf{Metric}
    & \textbf{Abn.}
    & \textbf{Fgn.}
    & \textbf{Loc.}
    & \textbf{Sev.}
    & \textbf{Trait}
    & \textbf{Trend}
    & \textbf{Desc.}
    & \textbf{Qual.}
    & \textbf{Rel.}
    & \textbf{Adv.} 
    & \\
    \midrule
    BERTScore
    & 96.99 & 90.99 & 91.98 & 96.98 & 96.83 & 96.49 & 94.90 & 88.72 & 80.25 & 91.0 & 92.53 \\
    \bottomrule
  \end{tabular}
  \normalsize
  \label{tab:check_extract}
\end{table}